%% file: main.tex
\documentclass[sigconf]{acmart}

\usepackage{comment}
\usepackage[normalem]{ulem}
\usepackage{subcaption}
\usepackage{nicefrac}
\usepackage{graphicx}
\usepackage{multirow}
\usepackage{colortbl}
\usepackage{rotating} 
\usepackage{longtable}
\usepackage{hyperref}
\usepackage[T1,T2A]{fontenc}
\usepackage{lmodern}

\usepackage{soul} 
\usepackage{ulem}
\usepackage{comment}
\renewcommand\footnotetextcopyrightpermission[1]{} % removes footnote with conference information in first column

\newcommand{\fixme}[1]{{\textcolor{red}{\em\bf{[FIXME: #1]}}}}
\AtBeginDocument{%
  \providecommand\BibTeX{{%
    \normalfont B\kern-0.5em{\scshape i\kern-0.25em b}\kern-0.8em\TeX}}}

\setcopyright{none}

\begin{document}

\title{What goes on inside rumour and non-rumour tweets and their reactions: A Psycholinguistic Analyses}


\author{Sabur Butt}
\authornote{Both authors contributed equally to this research.}
\email{sabur@nlp.cic.ipn.mx}
\orcid{0000-0002-4056-8923}
\affiliation[]{%
  \institution{Centro de Investigaci\'{o}n en Computaci\'{o}n del IPN}
  \city{Mexico city}
  \country{Mexico}
}

\author{Shakshi Sharma}
\authornotemark[1]
\email{shakshi.sharma@ut.ee}
\affiliation{%
  \institution{University of Tartu}
  \city{Tartu}
  \country{Estonia}
}

\author{Rajesh Sharma}
\email{rajesh.sharma@ut.ee}
\affiliation{%
  \institution{University of Tartu}
  \city{Tartu}
  \country{Estonia}
}

\author{Grigori Sidorov}
\email{sidorov@cic.ipn.mx}
\affiliation{%
  \institution{Centro de Investigaci\'{o}n en Computaci\'{o}n del IPN}
  \city{Mexico city}
  \country{Mexico}
}

\author{Alexander Gelbukh}
\email{gelbukh@cic.ipn.mx}
\affiliation{%
  \institution{Centro de Investigaci\'{o}n en Computaci\'{o}n del IPN}
  \city{Mexico city}
  \country{Mexico}
}

\pagestyle{plain} % removes running headers
\renewcommand{\shortauthors}{Sabur and Shakshi, et al.}

\begin{abstract}
In recent years, the problem of rumours on online social media (OSM) has attracted lots of attention. Researchers have started investigating from two main directions. First is the descriptive analysis of rumours and secondly, proposing techniques to detect (or classify) rumours. In the descriptive line of works, where researchers have tried to analyse rumours using NLP approaches, there isn't much emphasis on psycho-linguistics analyses of social media text. These kinds of analyses on rumour case studies are vital for drawing meaningful conclusions to mitigate misinformation. For our analysis, we explored the PHEME-9 rumour dataset (consisting of 9 events), including source tweets (both rumour and non-rumour categories) and response tweets. We compared the rumour and non-rumour source tweets and then their corresponding reply (response) tweets to understand how they differ linguistically for every incident. Furthermore, we also evaluated if these features can be used for classifying rumour vs. non-rumour tweets through machine learning models. To this end, we employed various classical and ensemble-based approaches. To filter out the highly discriminative psycholinguistic features, we explored the SHAP AI Explainability tool. To summarise, this research contributes by performing an in-depth psycholinguistic analysis of rumours related to various kinds of events.
\end{abstract}

\begin{CCSXML}
    <ccs2012>
    <concept>
        <concept_id>10002951.10003227.10003241.10003244</concept_id>
        <concept_desc>Information systems~Information systems applications</concept_desc>
        <concept_significance>500</concept_significance>
    </concept>
    <concept>
        <concept_id>10010147.10010257.10010258.10010259.10010264</concept_id>
        <concept_desc>Decision support systems~Data analytics</concept_desc>
        <concept_significance>300</concept_significance>
    </concept>
    </ccs2012>
\end{CCSXML}

\ccsdesc[500]{Information systems~Information systems applications}
\ccsdesc[300]{Decision support systems~Data analytics}

\keywords{Psycholinguistic Analyses, Rumour Detection, Explainable AI} 

\maketitle

\input{Introduction}
\input{RelatedWork}
\input{Dataset}

\input{Methodology}

\input{Analyses}

\input{Results}

\input{Conclusion}

\section*{Acknowledgment}
This work has received funding from the EU H2020 programm under the SoBigData++ project (grant agreement No. 871042), and from CHIST-ERA grant CHIST-ERA-19-XAI-010, by MUR (grant No. not yet available), FWF (grant No. I 5205), EPSRC (grant No. EP/V055712/1), NCN (grant No. 2020/02/Y/ST6/00064), ETAg (grant No. SLTAT21096), BNSF (grant No. КП-06-ДОО2/5). The work was done with partial support from the Mexican Government through the grant A1-S47854 of the CONACYT, Mexico and grants 20211784, 20211884, and 20211178 of the Secretar\'{i}a de Investigaci\'{o}n y Posgrado of the Instituto Polit\'{e}cnico Nacional, Mexico. The authors thank the CONACYT for the computing resources brought to them through the Plataforma de Aprendizaje Profundo para Tecnolog\'{i}as del Lenguaje of the Laboratorio de Superc\'{o}mputo of the INAOE, Mexico.

\bibliographystyle{ACM-Reference-Format}
\bibliography{main}
\label{sec:References}

\end{document}

%% file: Introduction.tex
\section{Introduction}

%Write a paragraph on Rumour detection
The credibility of information is the most decisive issue on social media as the unmoderated nature of social media text has resulted in several cases of misinformation spreading~\cite{derczynski2017semeval}. In this work, we focus on rumour, which is a specific kind of misinformation, whose authenticity has not been verified ~\cite{zubiaga2018detection}. In the past few years, researchers have analysed rumours from two different directions which can be divided into descriptive analyses of rumours, and the detection of rumours using a variety of machine learning and deep learning techniques ~\cite{ma2018rumor, geng2019rumor, sharma2021identifying}. %\fixme{@Shakshi: IJCNN citation}. 
Despite these apparent robust techniques, the increasing tendency to give rise to rumours motivates the development of systems that, by gathering and analysing the collective judgements of users~\cite{lukasik2016hawkes}, are able to reduce the spread of rumours by accelerating the sense-making process~\cite{derczynski2014pheme}. 

In particular, linguistics and natural language processing researchers have taken the onus to study how users have discussed rumours and to understand the psycho-linguistic attributes connected to rumour spreading and detection~\cite{ giachanou2021detection, shu2019defend}. Scientific studies aim to understand the malicious intentions of spreading rumours  through the psychological processes involved in the use of language can help in textual classification and behaviour analyses of users. Linguistics and natural language processing researchers study how users have discussed rumours and understand the psycho-linguistic attributes connected to rumour spreading and detection~\cite{ giachanou2021detection, shu2019defend}. See Section 2 for more details. 

To aid in the mitigation of misinformation, in this work, we performed the analysis of rumour vs. non-rumour tweets using psycholinguistic approaches, which is the study of the interrelation between linguistic factors and psychological aspects. It should be noted that we are not considering the user level features and are solely focusing on textual information. To be specific, we used psycholinguistics attributes that tend to convey the latent (hidden) meaning of the text. However, the patterns of these attributes cannot be determined in individual instances and need to have aggregated supervised data processed by computing psycho-linguistic methods and statistical evidence. To the best of our knowledge, this is the first study to use psycholinguistics features to conduct an in-depth analysis of rumour and non-rumour tweets.

We also investigated how effectively the characteristics of psycholinguistic analyses can assist classification algorithms for predicting rumour and non-rumour tweets. As a step further, we also looked at the ``why and what'' part. That is, to filter which features are more important in the identification of rumours. Specifically, we investigate the following research questions:

\textbf{RQ 1:} Is there a difference in psycho-linguistic characteristics between rumour and non-rumour source tweets?

\textbf{RQ 2:} How can psycho-linguistic features be used to differentiate between the reactions that rumour and non-rumour tweets attract?

\textbf{RQ 3:} Does the contribution of psycho-linguistic features vary from event to event or do they remain consistent for all events?

\textbf{RQ 4:} Can we exploit these features for classifying rumour and non-rumour tweets using Machine Learning models?

\textbf{RQ 5:} Which psycho-linguistic features are highly discriminative for identifying rumour and non-rumour class for the classification task? 

To answer these research questions, we used the PHEME-9 dataset, consisting of 9 different events (Section 3). We performed psycho-linguistic studies to address RQ1, RQ2 and RQ3, and extracted four types of features from the dataset: LIWC, Readability, SenticNet, and Emotions. All of these features provided us with more insight and perspective into the rumour and non-rumour tweets. We calculated the statistical significance and mean values of every psycho-linguistic feature for rumour source tweets, non-rumour source tweets, reactions of rumour tweets, and reactions of non-rumour tweets, respectively. The statistical significance tests helped us to assess the difference between the rumour and non-rumour psycho-linguistic features. We explain the methodology in Section 4 and the results of our analysis in Section 5. For RQ4 and RQ5, we further used machine learning classifiers (classical as well as ensemble-based) on every event to evaluate the effectiveness of the psycholinguistic features in classifying the tweets into rumour vs. non-rumour classes. Lastly, we use SHAP, an AI Explainability tool based on shapley values to identify the measure of contributions each feature has in the model (Section 6). Finally, we conclude this article with some future works in Section \ref{sec:conclusion}.

%% file: RelatedWork.tex
\section{Related Work}

Psychological perspectives have been used in several studies to find a correlation with conspiracy theories. Douglas et al.~\cite{douglas2017psychology} wrote extensively on the psychological factors and divided them into existential (e.g. desire for control), social (e.g. desire to maintain a positive image of the self or group) and epistemic (e.g. desire for understanding) factors. The author explained that these factors contribute to the popularity of conspiracy theories. Another study discovered~\cite{lantian2017know} that conspiracies urge from the need for uniqueness. Similarly, in the light of the COVID-19 pandemic, researchers~\cite{callaghan2019parent} have correlated a high level of conspiracy thinking of parents with the delay of vaccination among children. While there are studies that have attempted to discover the correlation of psycholinguistics with fake news~\cite{shu2019defend} and conspiracy~\cite{rains2021psycholinguistic}  in general, we narrowed it down to the understanding of rumour. Various NLP studies have been published related to rumour detection in text. %\fixme{Please check next 2 lines.}
These  studies mostly revolve around the data collection, techniques and features suitable for identifying rumour text. Other than PHEME which is the benchmark dataset for rumour detection and is used for this study, the popular text based datasets include RUMDECT~\cite{ma2016detecting}, SNAP data~\cite{yang2011patterns}, CrisisLexT26~\cite{olteanu2015expect}, MULTI~\cite{jin2017multimodal}, KWON~\cite{kwon2013prominent} and RUMOUREVAL~\cite{derczynski2017semeval}. 

Among the components of rumour detection, rumour has been explored~\cite{varshney2020review} in the context of rumour tracking, rumour stance classification, rumour detection, and rumour veracity classification. We analyzed the most commonly used techniques for rumour tasks and observed the use of traditional Machine learning (TML) methods with selective feature engineering, deep learning models (DL) and hybrid models respectively. In the study~\cite{zubiaga2016learning} authors introduced context-aware rumour detection using a sequential classifier to detect rumours from the tweets divided into five news events. In an outbreak of stories, identifying the emergency is also important for the timely detection of credible information. The work~\cite{xia2012information} used an unsupervised algorithm to label the tweets as credible and incredible and identified the urgency of the news verification using supervised machine learning methods and multiple features (content, author, and diffusion). The Rumour detection task can be topic-level~\cite{yang2015emerging} or post-level~\cite{derczynski2017semeval}, where the task is to identify if the topic is relevant to the text or if the post information has rumour respectively. The trend of rumour detection on topic-level has shifted quickly towards the post level understanding of rumour and, hence, needs insightful analyses for understanding linguistic and psycho-linguistic attributes of rumour. 

The need to understand the explainability aspect of rumours stems from the need for early detection. Early detection of the rumour is essential to mitigate the harm and the study~\cite{ma2018rumor} segregated the rumours using a propagation tree. They used recursive neural networks to classify the tweets into false rumour, non-rumour, unverified rumour and true rumour. A Recurrent Neural Network~(RNN) based aware provenance approach was proposed~\cite{duong2017provenance} combining the textual information and provenance information to enhance the results. To tackle the cases where provenance details were missing, they used a fusion of text and provenance information. Among the machine learning models~\cite{ashraf2021cic, geng2019rumor, jin2017multimodal, sampson2016leveraging} Support vector machine (SVM), Naive Bayes (NB),
Random Forest (RF), Decision Tree (DT), Logistic Regression (LR), and XGBoost have been used repeatedly. Deep learning models (Convolutional neural network (CNN), Long short-term memory (LSTM), Recurrent neural networks (RNN))~\cite{santhoshkumar2020earlier, kotteti2020ensemble, nguyen2017comprehensive} and transformer-based models (XLNet, BERT, RoBERTa, DistilBERT, and ALBERT)~\cite{transformer1, transformer2} have often given the state of the art results in multiple rumour tasks.

Psycholinguistic features on rumour detection studies have involved the use of Long Short-Term Memory (LSTM) with LIWC features i.e swear words and personal pronouns~\cite{tausczik2010psychological} and emotions~\cite{vosoughi2018spread}. The study~\cite{vosoughi2018spread} showed the existence of false rumours that triggered fear, disgust and surprise in the rumour replies, while,~\cite{giachanou2019leveraging} proposed an LSTM-based model with emotions to check the credibility of articles. The role of user profiles was evaluated by ~\cite{shu2018understanding} which showed implicit (age, location) and explicit features (follower count, status count) that contributed to fake news.  In addition, they showed how the combination of these features with the psycho-linguistic characteristics (LIWC, Writing styles) can be effective. Another study~\cite{giachanou2020role} proposed a CNN based model combined with personality traits (Big-five) and LIWC to distinguish between fake users.  Similarly, the authors~\cite{giachanou2021detection} gave a comparative analysis over various profile, linguistic and psychological characteristics (Big-five personality traits, LIWC, emotions) and used CNN based classifier with word embeddings and psycho-linguistic characteristics for classification. 

Contrary to these studies, our main focus is to
provide insights into the source as well as reaction tweets comparing different events as every event triggers a different insight. Moreover, we discuss the explainability part in the context of which of these features highly contribute to the rumour or non-rumour classes as well as their reactions. It should be noted that none of these papers indulges in the explainability aspects of psycholinguistics in rumour tweets and reaction tweets.

%% file: Dataset.tex
\section{Dataset}

This study has been carried out using the PHEME dataset~\cite{kochkina_liakata_zubiaga_2018}, which consists of nine events, wherein all the events are breaking news. To be specific, each event contains the source tweets which are divided into rumour and non-rumour source tweets. Similarly, every source tweet has a set of reaction tweets which are again divided into rumoured reaction tweets and non-rumour reaction tweets. The reactions are triggered by either a rumoured source tweets and the non-rumoured source tweets and the reactions triggered by rumour and non-rumour tweets are set into rumoured reaction tweets and non-rumoured reaction tweets, respectively. However, it should be noted that the reactions have no ground truth to be labelled as the rumour or non-rumour and are just a reaction in the form of replies. The dataset initially reported five events~\cite{zubiaga2016learning}, but was later extended to nine events to bring more variety in context. Understanding the events for rumours in the text also helps us understand the linguistic differences since these incidences were rife with rumours and gained significant media attention. The nine incidences mentioned are explained below:

\begin{itemize}
    \item Ferguson unrest: The incident refers to the protest in Michigan, USA after an 18 year old African-American was shot by a white police officer. 
    \item Ottawa shooting: The incident occurred in Canada's Ottawa Parliament Hill, which resulted in the death of a Canadian soldier.
    \item Sydney siege: Lindt cafe siege in Sydney was a terrorist attack when a gunman held several people, hostage, in a cafe. 
    \item Charlie Hebdo shooting: The weekly Charlie Hebdo satirical newspaper office was invaded by two brothers in Paris, which resulted in 12 people being killed and 11 injured. 
    \item Germanwings plane crash: The plane enroute from Barcelona to Dsseldorf was crashed due to the suicidal tendencies of the co-pilot. All passengers and crew were killed in the crash. 
    \item Gurlitt: The incident relates to the Gurlitt art collection. Hildebrand Gurlitt, was an art merchant who acted for the Nazis, who died with no direct descendants. People started spreading rumours about the acceptance of the artwork by Berns Museum. 
    \item Prince-Toronto: The rumour started with a deleted tweet of a "pop-up show" by the Prince's band. People suspected a surprise secret concert. 
    \item Putin missing: The incident refers to the public disappearance of the Russian president. 
    \item Ebola: The soccer star Michael Essien was thought to have contracted Ebola and the rumour went viral.  
\end{itemize}

Table~\ref{datastatspheme9} shows the statistical distribution among source and reaction tweets. The Ebola case study had no list of rumour tweets, so we disregarded the case study and performed the experiments with the remaining eight case studies. 
%\fixme{Are we using all the 9 incidents in our analysis ?}

\begin{table}[!hbt]
\caption{Data distribution of PHEME-9 among non-rumour and rumour tweets in source and reactions based on each individual case. NR and R represents non-rumour and rumour respectively}
  \label{datastatspheme9}
\begin{tabular}{l|l|l|l|l|}
\cline{2-5}
\multicolumn{1}{c|}{\textbf{}}          & \multicolumn{2}{c|}{\textbf{Source tweets}}                        & \multicolumn{2}{c|}{\textbf{Reaction tweets}}                      \\ \cline{2-5} 
                                        & \multicolumn{1}{c|}{\textbf{NR}} & \multicolumn{1}{c|}{\textbf{R}} & \multicolumn{1}{c|}{\textbf{NR}} & \multicolumn{1}{c|}{\textbf{R}} \\ \hline
\multicolumn{1}{|l|}{\textbf{Charlie}}  &            1621                      &           458                      &             29302                     &    68887                             \\ \hline
\multicolumn{1}{|l|}{\textbf{German}}   &            231                   &               238                  &            1764                      &           2256                    \\ \hline
\multicolumn{1}{|l|}{\textbf{Sydney}}   &            699                    &              522                    &          14621                       &         8154                        \\ \hline
\multicolumn{1}{|l|}{\textbf{Putin}}    &            112                      &             126                    &        236                      &              361                   \\ \hline
\multicolumn{1}{|l|}{\textbf{Prince}}   &             4                     &               229                  &          3                        &              666                   \\ \hline
\multicolumn{1}{|l|}{\textbf{Ottawa}}   &          420                        &              470                   &         5428                         &         5966                        \\ \hline
\multicolumn{1}{|l|}{\textbf{Gurlitt}}  &            77                      &                61                 &          15                        &         26                       \\ \hline
\multicolumn{1}{|l|}{\textbf{Ferguson}} &            859                      &             284                    &        16837                          &        6195                         \\ \hline
\end{tabular}
\end{table}

%% file: Methodology.tex
\section{Methodology}
We combined various psycholinguistic features such as LIWC, SenticNet~\cite{cambria2014sentic}, readability indexes, and emotions to bring out true insights about user patterns. In computerized text analyses, Linguistic Inquiry and Word Count~\cite{pennebaker2015development} is a gold standard in understanding linguistic aspects of motivations, thoughts, feelings and personality. SenticNet is used to derive concept-level sentiment analyses from the text. Readability features indicate the easiness to interpret a text depending on its unique attributes. Finally, emotions show us the true nature of the feelings which cannot be interpreted by plain polarity and sentiment detection. 

%\fixme{@Sabur: If LIWC is a part/kind of psycholinguistic featuers? If yes, then the next part of line appears as if we are combining LIWC + psycholingusitic features such as SentiNet, etc ..} 

\subsection{Pre-processing and Feature extraction}\label{sec:feature}
We first thoroughly pre-processed the noisy data of Twitter and then extracted previously mentioned features from it. The LIWC features were extracted through the LIWC~(2015) program which records all punctuations and words. We used Textatistic~\footnote{\url{https://pypi.org/project/textatistic/}} python library to extract readability features including Gunning Fog, Flesch Reading Ease, Flesch-Kincaid, Simple Measure of Gobbledygook (SMOG) and Dale-Chall. It is important to keep the period in a sentence to extract readability features, hence, we only removed hashtags, user mentions, emojis and URL's. SenticNet features were calculated using the SenticNet API~\footnote{\url{https://sentic.net/api}} and the sentic words were combined to achieve the phrase features. For SenticNet features, we used word lemmatization and removed all punctuations, URL's, user-mentions, hashtags, and custom stopwords (without negation words) for pre-processing. Emotions were calculated using DistilRoBERTa base model~\cite{Sanh2019DistilBERTAD} trained on a combination of multiple emotion datasets for English predicting Ekman's 6 basic emotions, plus a neutral class which can be tested on the HuggingFace API~\footnote{\url{https://huggingface.co/j-hartmann/emotion-english-distilroberta-base?}}. 

\subsection{Statistical Test and Machine Learning}
%#Which statistical method was used for statistical evidence
%\textbf{- Statistical test - ML methods (decision tree, random forest, and SVM))} To be added by Shakshi....
To verify that the difference between the features (indicated in Section \ref{sec:feature}) of rumour and non-rumour classes are significant, we employed Kolmogorov Smirnov (KS) test\footnote{https://en.wikipedia.org/wiki/Kolmogorov\%E2\%80\%93Smirnov\_test}. This is a non-parametric test used to compare the two distributions. In our case, the two distributions correspond to rumour and non-rumour features, as explained in Section \ref{sec:analyses}. 

Next in Section \ref{sec:results}, we used these features as inputs to machine learning models to evaluate whether they are good enough to classify the tweet as a rumour or non-rumour. Furthermore, we utilized the SHAP explainability tool\footnote{https://shap.readthedocs.io/en/latest/overviews.html}, indicating which features are more relevant than others when making predictions.

%% file: Analyses.tex
\section{Analyses}\label{sec:analyses}

In this section, using psycholinguistic analysis we studied the characteristics of rumour and non-rumour source tweets and the corresponding reactions to the tweets. Different events show us insightful information on the nature of rumour categories and how the social and psychological meaning of words can change in every scenario. In addition, we calculated the statistical significance test to validate that the difference in both classes (rumour and non-rumour) was not due to random chance. We used the mean value to evaluate the overall influence of the individual features for every class. 

%The section presents the complete analyses of all psycholinguistic features used for psychological understanding of the tweets. 
%of users who produced the tweets. 

%\sabur{The statistical significance tells us that the likelihood of the difference in both classes (rumour and non-rumour) are not due to random chance. The mean value helps us to evaluate the overall influence of the individual features on every class.} 

\subsection{LIWC}
LIWC features tell us about the psychometric properties of the tweets. Table~\ref{tab:LIWCsourcetweets} and~\ref{tab:LIWCreactiontweets} shows the results of individual events for source tweets and reaction tweets respectively. The Tables clearly show a significant difference between rumour and non-rumour tweets based on LIWC categories. The Table~\ref{tab:LIWCaggregatedstats} on the other hand is an aggregated representation of the significant differences for LIWC features in source tweets and their reactions. %\sak{\sout{Should we also mention here about Table no 2? and also, should we talk about why we have two types of tables - table 2 which is about aggregated and table 3,4 which is about individual events?}}

\subsubsection{Linguistic Processes} 
These processes include text related analysis. In particular,
\textbf{Word Count} (WC) tells us about the engagement and domination of a user in a conversation. Word count needs to be balanced in the deceptive scenarios where the descriptiveness of the scenario needs to be balanced and too many words can reveal inaccuracies.  In every event, word count was proven to be statistically significant in either the source tweets or the reaction tweets. The average word count (AWC) of all events of rumour source tweets was 20 and the average word count for the rumoured reply tweets was 14.63.
\textit{Where the average word count of rumoured source tweets and non-rumoured source tweets was almost the same, AWC in non-rumoured reply tweets was found to be higher than rumoured reply tweets. One possible reason for this can be that source tweets needed more convincing and engagement and also engaged more non-rumoured reply tweets to deny the rumoured claims}. %\fixme{\sout{Raj: Sounds like causality? We need to be careful. The previous line sounds more like causality but we need to write -- One possible reason......}}. 
The \textbf{function words} (Table~\ref{tab:LIWCsourcetweets} and~\ref{tab:LIWCreactiontweets}, Row 2) in LIWC include total pronouns, impersonal pronouns, articles, prepositions, auxiliary verbs, common adverbs, conjunctions and negations. Among the personal pronouns (I, s/he, they, we), we can identify that the mean in rumour source tweets is 1 against the 3.20 in non-rumour source tweets.  Pronouns give us a lot of insight into the personality of the users as they indicate how users are communicating with each other\cite{baddeley2008telling, arguello2006talk, simmons2008hostile} and what is the intent~\cite{berlyne1960conflict} of the conversation. \textit{We saw a means score of 1st person singular (I), 1st person plural (we), 2nd person(you), 3rd person singular (S/he) and third-person plural (they) all to be higher in non-rumour source tweets. The same trend was seen in the rumour and non-rumour reaction tweets where, the use of personal pronouns was significantly more however, the non-rumour personal pronouns still had a higher mean.} Although the collective significance of function words in reaction tweets was not seen, when we divided the significance based on events (\ref{tab:LIWCreactiontweets}), we saw 5 out of 8 events showing significant difference between non-rumour and rumour reaction tweets. \textit{Use of \textbf{prepositions} (to, with above) shows us concern with precision~\cite{pennebaker2003words, newman2003lying} and was found to be higher in rumour source tweets ($\mu$ = 11) compared to rumour reaction tweets ($\mu$ = 7.81)}. \textbf{Negation words} psychologically correlate with inhibition~\cite{taylor2008linguistic, pennebaker1997linguistic} and was seen to be higher in rumour reaction tweets with the mean value of 2.02 compared to 1 in rumour source tweets. \textit{In both events, non-rumour tweets has higher negation words than rumour tweets. Similarly, other attributes such as \textbf{conjunction} ($\mu$ = 2 in RS, $\mu$ = 2.39 in NRS), \textbf{common adverbs} ($\mu$ = 2 in RS, $\mu$ = 2.35 in NRS), \textbf{auxiliary verbs} ($\mu$ = 4 in RS, $\mu$ = 4.71 in NRS) and \textbf{impersonal pronouns} ($\mu$ = 1 in RS, $\mu$ = 2.34 in NRS) all saw lower mean in rumour source tweets. Use of Impersonal pronouns, auxiliary verbs, conjunctions and negation words was seen to be more in reaction tweets where non-rumour reactions had a higher mean than rumour reactions}.

\subsubsection{Psychological Processes} Among the psychological processes we analyzed \textbf{affective process} (positive emotions, negative emotions), \textbf{social processes} (female references, male references, family, friends), \textbf{cognitive processes} (insight, causation, discrepancy, tentative, certainty, differentiation), \textbf{perceptual processes} (see, hear, feel), \textbf{biological processes} (body, sexual, ingestion, health), \textbf{drives} (affiliation, achievement, power, reward, risk), \textbf{time orientations} (past focus, future focus, present focus), \textbf{relativity} (motion, space, time) and \textbf{personal concerns} (work, leisure, home, money, religion). %\fixme{\sout{should we break the para from here? If yes, then make "affective processes" bold.}} 

Though \textbf{affective processes} are discussed extensively in Section~\ref{EmotionAnalyses}, \textit{LIWC stats showed positive emotions words ($\mu$ = 1 in RS, $\mu$ = 2.16 in NRS) and negative emotions ($\mu$ = 2 in RS, $\mu$ = 3.19 in NRS) breaking down to more anxiety, anger and sadness in the non-rumour source tweets.
The rumour ($\mu$ = 7.034) and non-rumour ($\mu$ = 7.624) reaction stats show us a similar story with higher affective processes in non-rumour tweets}. One can observe (Table~\ref{tab:LIWCsourcetweets} and~\ref{tab:LIWCreactiontweets}, Row 3) a huge disparity between the positive and negative mean of words in the rumour source tweets showing the imbalance of emotions. On the other hand, reaction tweets show a very similar mean %\fixme{not sure where "mean" comes first, but we need to justify - why we used mean. If this is the first place, we can make a footnote?}
of negative words as reactions to rumours are meant to trigger negative emotions.  

\textbf{Social words} correlate with social concerns~\cite{newman2003lying, newman2008gender, tausczik2010psychological}, and social support~\cite{leshed2007feedback}, we observed (Table~\ref{tab:LIWCsourcetweets} and~\ref{tab:LIWCreactiontweets}, Row 4) a variety of observations in both source and reaction tweets. Type of rumour can draw many such references and one can see a pattern of more female ($\mu$ = 0.0598 in RS, $\mu$ = 0.1196 in NRS) and friends ($\mu$ = 0.0695 in RS, $\mu$ = 0.1465 in NRS) reference words in non-rumour source tweets and more family ($\mu$ = 0.2274 in RS, $\mu$ = 0.1493 in NRS) and male ($\mu$ = 0.4901 in RS, $\mu$ = 0.3109 in NRS) average words in rumour source tweets.

\textbf{Cognitive Processes} gives us the insight of the reasoning and difference in the thought process of the authors~\cite{arguello2006talk, batten2002physical, liehr2002expressing, tausczik2010psychological}. Cognitive processes showed a significant difference in the events of both reaction and source tweets (Table~\ref{tab:LIWCsourcetweets} and~\ref{tab:LIWCreactiontweets}, Row 5) %\sak{\sout{I think this is about the Table no 2? So, we should cite it?}}.
The event of Putin and Prince had no significant difference in cognitive processes due to the nature of the rumour (no supporting claim). \textit{Cognitive processes were also used higher by the users in non-rumour source and reaction tweets. Except for the tentative words (maybe, perhaps) that were used more in the rumour source tweets and reaction tweets, all subcategories of cognitive processes had a higher value of mean in the non-rumour source and reaction tweets}. 

\textbf{Perceptual Processes} tell us about the sensory experiences in the text including seeing, hearing and feeling related words. \textit{We saw that the perceptual processes (Table~\ref{tab:LIWCsourcetweets} and~\ref{tab:LIWCreactiontweets}, Row 6) did not create a lot of impact in the individual scenarios except in Putin's case in the source tweets where the category was very relevant to the scenario of Putin being absent since his last sighting}. Similarly, \textbf{biological processes} (Body, health, sexual, ingestion) was also very scenario specific on significant in Sydney's case was about a hostage scenario and an act of terrorism. \textit{We conclude that biological processes (Table~\ref{tab:LIWCsourcetweets} and~\ref{tab:LIWCreactiontweets}, Row 7) become significant where the incident is directly related to health, body, sex etc}. %\sak{\sout{The lines of biological process seems unclear. I mean what is the conclusion or what are we trying to say here?}}

\textbf{Time orientations} in the psychological processes gave us good insight about the rumour and non-rumour scenarios where \textit{rumour source tweets (Table~\ref{tab:LIWCsourcetweets}, Row 13) were more past focused ($\mu$ = 3 in RS, $\mu$ = 1.52 in NRS) and non-rumour source tweets were more present focused ($\mu$ = 6.42 in NRS, $\mu$ = 5 in RS)}. Reactions (Table~\ref{tab:LIWCreactiontweets}, Row 13) to non-rumour and rumour tweets followed the same trend.  

\textbf{Relativity} (area, bend, exit) includes motion (arrive, car, go), space (down, in, thin), time (end, until, season) related words. Relativity, in general, proved to be less significant in reactions (Table~\ref{tab:LIWCreactiontweets}, Row 9) of the scenarios and significant in five out of eight scenarios in the source tweets (Table~\ref{tab:LIWCsourcetweets}, Row 9). \textit{Though relativity is more heavily used in rumour source tweets and their reactions, however, an important observation to make is that motion related words were present more in the non-rumour source tweets and the reactions of the rumour tweets. A combination of higher mean in the negation words and motion related words in the non-rumour source tweets shows us the attempted correction in the direction of the conversation}. 

People show their personal concerns in the reactions and sources of the tweet where we can see an emphasis on work, money, leisure, home religion and death related words. \textbf{Personal concerns} can be seen to be significant in some events (Table~\ref{tab:LIWCsourcetweets} and~\ref{tab:LIWCreactiontweets}, Row 12) i.e Charlie, Sydney and Ferguson in the source tweets. However, this can be case dependent as all these events were linked to violence and reported abuse. \textit{In the incidences that mattered, we saw a higher mean of work, death and leisure related words in the rumoured cases whereas more use of home, money and religion related words in the non-rumoured source tweets}. \textbf{Informal Language} (Table~\ref{tab:LIWCsourcetweets} and~\ref{tab:LIWCreactiontweets}, Row 10) on social media is expected in general and the data presented a similar scenario where the \textit{mean of non-rumoured source tweet ($\mu$ = 5.32 in NRS, $\mu$ = 5 in RS) was higher in mean}. In general, the reaction of the tweets had a high mean ($\mu$ = 4.024 in NRS, $\mu$ = 4.051 in RS) of informal words as well.

\textbf{Drives} (Table~\ref{tab:LIWCsourcetweets} and~\ref{tab:LIWCreactiontweets}, Row 8) are the motivational factors and in rumour detection it can identify a lot about the authors motives/agendas behind the tweets. Drives might include affiliation (ally, friend), power (superior, bully), reward (prize, benefit), risk (danger, doubt), and achievement (win, success) related words. Drives had a significant impact on the differentiation between non-rumour and rumour resources. Breaking down the reaction on rumour and non-rumour tweets, it also played a vital role in differentiating how drives of users were different in every scenario impacting through the rumour and non-rumour sources. Further narrowing revealed that \textit{rumoured sources had higher mean in reward, risk and power related words and non-rumour source tweets had more mean of affiliation and achievement related words}. 

\subsubsection{Punctuation}
Punctuations (Table~\ref{tab:LIWCsourcetweets} and~\ref{tab:LIWCreactiontweets}, Row 11) such as question marks and assents show how people are communicating with each other. Punctuations can also show us attempted effort for explanation or emphasis through indicators such as apostrophes and parentheses. \textit{Punctuations did not have a lot of impact in the reactions of rumour and non-rumour, however, played some part in differentiating the rumour and non-rumour source tweets. We observed that non-rumour sources of tweets have relatively higher mean periods, commas, semi-colons, question marks, apostrophes and parenthesis}. The reactions, in general, had more mean of punctuations for both rumour and non-rumour. 

\subsubsection{Other Grammar}

The other grammar category (Table~\ref{tab:LIWCsourcetweets} and~\ref{tab:LIWCreactiontweets}, Row 14) is the only category that is statistically significant in both rumour and non-rumour source tweets and their reactions (Table~\ref{tab:LIWCaggregatedstats}, Row 14). The grammar category of LIWC includes common verbs, common adjectives, comparisons, interrogatives and quantifiers. Common verbs can explain the temporal focus of the tweets along with common adjectives that identify the actions. In the non-rumour and rumour source tweets, common adjectives almost show similar means and hence do not contribute to differentiating. \textit{The reactions to the rumour and non-rumour tweets however used much more common adjectives and verbs. Non-rumour source tweets and the reactions to non-rumour source tweets had a higher mean of interrogative and comparison words compared to rumour source tweets and their reactions, signifying that users were questioning more about the non-rumour tweets compared to rumour tweets. Reaction of both categories also showed a greater mean of quantifiers}.

\input{Tables/LIWC_statspercase}

\subsection{Readability}
Readability allows differentiating between a text that is easy to comprehend compared to a text that is complicated and requires a high level of education or intelligence for understanding. There are many readability scores used to evaluate the text, we considered the most popular tests to evaluate tweets. Table~\ref{tab:Readabilitysourcetweets} and~\ref{tab:Readabilityreactiontweets} shows the significant difference in readability scores between rumour and non-rumour tweets and their reactions to various events. The Table presents Flesch score~\cite{farr1951simplification}, Flesch-Kincaid score~\cite{kincaid1975derivation}, Gunning-Fog score~\cite{gunning1969fog}, Smog score~\cite{mc1969smog}, Dale-Chall score~\cite{flesch1948new}. The Flesch score in source tweets ($\mu$ = 72.26  in RS, $\mu$ = 74.12 in NRS) indicated "fairly easy to use" text and reactions ($\mu$ = 80.04  in R, $\mu$ = 79.57 in NR) indicated the same trend, where the high number of the score (0-100) indicates more easiness to read. If we translate that score to US academic grade level (0-18) using Flesch-Kincaid we see that the source tweets ($\mu$ = 7.22 in RS, $\mu$ = 7.76 in NRS) and their reactions ($\mu$ = 4.66 in R, $\mu$ = 4.89 in NR) show a 7th grade of understanding to read the text. The Gunning-Fog score (grades 1-20) points the source tweets ($\mu$ = 8.9  in RS, $\mu$ = 9.41 in NRS) and their reactions ($\mu$ = 7.2 in R, $\mu$ = 7.51 in NR) in the direction of Smog score ($\mu$ = 7.6 in RS, $\mu$ = 8.11 in NRS) where the easiness of reading is suitable for a 7th-9th grader. Lastly, Dale-Chall score for readability indicated the text required the reading comprehension of 11th to 15th grade student. Although we found no significant difference collectively of readability scores, however, we found many incidences of significance individually when divided per event. 
\textit{It would be fair to say that Twitter rumour spreaders and their responses engage casual conversations and are designed to target masses for rapid spreading of news which is different from formal news platforms}. 

\input{Tables/Readability_stats}

\subsection{SenticNet}
SenticNet features $\in [-1,+1]$ give us a commonsense understanding of the text by translating the hourglass wheel of emotions~\cite{plutchik2001nature, cambria2012hourglass} into statistical values. We considered the sentic values (Aptitude, Pleasantness, Attention, and Sensitivity) and polarity associated with the concept. We observed a shift of emotions throughout the tweets giving a mix of SenticNet values. The aggregated statistical significance can be seen in Table~\ref{tab:SenticAggregate} which shows all SenticNet values significant for rumour and non-rumour source tweets and their reactions. \textit{Rumour source tweets were weighted more towards pleasantness ($\mu$ = 0.07872 in RS, $\mu$ = 0.03275 in NRS), attention ($\mu$ = 0.11747 in RS, $\mu$ = 0.08301 in NRS) and polarity ($\mu$ = 0.10902 in RS, $\mu$ = 0.08646 in NRS) while non-rumour tweets had more mean of sensitivity ($\mu$ = 0.03 in RS, $\mu$ = 0.05351 in NRS) related emotions}. The mean value of aptitude ($\mu$ = 0.09925 in RS, $\mu$ = 0.09879 in NRS) on the other hand can be seen to be very close although towards positive aptitude emotions (trust and acceptance). Among the reaction tweets we saw no aggregate statistical significance for any of SenticNet values, however,  Table~\ref{tab:SenticNetreaction} shows significance in many individual scenarios such as the Sydney case where all SenticNet features had significant differences among reactions. Reactions of rumour and non-rumour tweets gave us negative mean pleasantness ($\mu$ = -0.01907 in R, $\mu$ = -0.0258 in NR) weighing more towards non-rumour tweets along with sensitivity ($\mu$ = 0.0756 in R, $\mu$ = 0.0853 in NR), polarity ($\mu$ = 0.0373 in R, $\mu$ = 0.0383 in NR) and aptitude ($\mu$ = 0.05901 in R, $\mu$ = 0.0681 in NR). \textit{Attention related emotions (interest, anticipation and vigilance) were seen more in rumour reactions ($\mu$ = 0.04763 in R, $\mu$ = 0.0389 in NR) clearly showing how people are more attentive towards rumoured content that is designed to draw more attention}. The statistics identify that Sentic net values for non-rumour reactions draw more attention to emotions like annoyance, anger, acceptance, trust, grief and sadness compared to rumour reactions. It should be noted that the emotions are triggered depending on the scenario and drive of rumour spreaders. Table~\ref{tab:SenticNetsource} clearly shows a significant difference of SenticNet values in source tweets showing incidences like Gurliit in which sentic values had no significance and Sydneys case where sentic values plays important role in differentiation rumour and non-rumour sources.

\input{Tables/SenticNet_stats}

\subsection{Emotions}~\label{EmotionAnalyses}

Table~\ref{tab:Emotiondistribution} shows the emotion percentages across the rumour and non-rumour categories. \textit{Fear and Sadness are the two most instigated emotions in the rumour tweets. The reactions to the rumour showed that the percentage of fear and sadness was converted into anger and surprise while the highest instigated emotion being neutral. Non-rumour source tweets had the highest percentage of neutrality, followed by fear, anger and sadness. The reactions to non-rumour source tweets had less percentage of fear and greater percentages of neutrality, anger and surprise}. We can see the patterns of rumoured tweets trying to use negative emotions to instil fear among people and as a reaction, many people felt fear compared to reactions to non-rumour sources in general. It should be noted that the majority of the rumoured incidences in the study were related to some sort of tragedy, however, the distribution among the same news in rumoured and non-rumoured forms shows the extent of negative and positive emotions used to achieve the potential motives.  

\input{Tables/Emotion_stats}

\begin{table*}[!htbp]
\centering
\caption{The table shows the partition of features according to categories and \textcolor{blue}{higher mean}. PP and CP denotes Personal Pronouns and Cognitive Processes, respectively}
\label{tbl:featurepartition}
\begin{tabular}{|c|l|l|}
\hline
\textbf{} & \textbf{Higher mean in source tweets} & \textbf{Higher mean in source tweets reply tweets} \\ \hline
\textbf{Rumour} & \begin{tabular}[c]{@{}l@{}}-Prepositions\\ -Social words (family)\\ -Cognitive Processes (tentative)\\ -Time (past)\\ -Relativity\\ -Personal (work, death, leisure)\\ -Drives (reward, risk, power)\\ -SenticNet (pleasantness, attention, polarity)\\ -Emotions (fear, sadness)\end{tabular} & \begin{tabular}[c]{@{}l@{}}-Negation words\\ -Cognitive Processes (tentative)\\ -Time (past)\\ -Relativity\\ -Emotions (anger, surprise)\end{tabular} \\ \hline
\textbf{Non-Rumour} & \begin{tabular}[c]{@{}l@{}}-Average word count\\ -PP (1st person {[}singular (I), plural (we){]}, 2nd person (you), \\   3rd person {[}singular (S/he), plural (they){]})\\ -Conjunction, common adverbs, auxiliary verbs\\ -Affective processes\\ -CP (insight, causation, discrepancy, certainty, differentiation)\\ -Time (present)\\ -Personal (home, money, religion)\\ -Informal language\\ -Drives (affiliation, achievement)\\ -Grammar (interrogative, comparison)\\ -SenticNet (sensitivity)\\ -Emotions (neutral, fear, anger)\end{tabular} & \begin{tabular}[c]{@{}l@{}}-Negation words\\ -Affective processes\\ -CP (insight, causation, discrepancy, \\   certainty, differentiation)\\ -Impersonal pronouns, auxiliary verbs, conjunctions\\ -Time (present)\\ -Grammar (interrogative, comparison)\\ -SenticNet (pleasantness, sensitivity)\\ -Emotions (fear)\\ -PP\end{tabular} \\ \hline
\end{tabular}
\end{table*}

The Table~\ref{tbl:featurepartition} gives a bird's eye view to the analyses section. We partitioned the rumour and non-rumour source tweets and their reactions based on feature impact and mean. The analysis is further strengthened by the SHAP plots and explanation given in Section~\ref{sec:results}.

%% file: Tables/LIWC_statspercase.tex
\begin{table}[!htbp]
  \caption{The table shows the significant difference of all \textcolor{blue}{LIWC} features between the non-rumour and rumour \textcolor{blue}{source tweets}}
  \label{tab:LIWCsourcetweets}
  \resizebox{\columnwidth}{!}
  {%
\begin{tabular}{ccclclclclclclcl}
\textbf{}                      & \textbf{Charlie}                 & \multicolumn{2}{c}{\textbf{German}}                  & \multicolumn{2}{c}{\textbf{Sydney}}                  & \multicolumn{2}{c}{\textbf{Putin}}                 & \multicolumn{2}{c}{\textbf{Prince}}                  & \multicolumn{2}{c}{\textbf{Ottawa}}                  & \multicolumn{2}{c}{\textbf{Gurliit}}                 & \multicolumn{2}{c}{\textbf{Ferguson}}                 \\
\textbf{WC}                    & \cellcolor[HTML]{EA9999}0.95     & \multicolumn{2}{c}{\cellcolor[HTML]{B6D7A8}0.006}    & \multicolumn{2}{c}{\cellcolor[HTML]{EA9999}0.98}     & \multicolumn{2}{c}{\cellcolor[HTML]{EA9999}0.129}  & \multicolumn{2}{c}{\cellcolor[HTML]{B6D7A8}0.04}     & \multicolumn{2}{c}{\cellcolor[HTML]{B6D7A8}7.02E-05} & \multicolumn{2}{c}{\cellcolor[HTML]{B6D7A8}0.0002}   & \multicolumn{2}{c}{\cellcolor[HTML]{B6D7A8}8.04E-05}  \\
\textbf{Function words}        & \cellcolor[HTML]{B6D7A8}2.44E-15 & \multicolumn{2}{c}{\cellcolor[HTML]{EA9999}0.07}     & \multicolumn{2}{c}{\cellcolor[HTML]{B6D7A8}7.45E-18} & \multicolumn{2}{c}{\cellcolor[HTML]{B6D7A8}0.012}  & \multicolumn{2}{c}{\cellcolor[HTML]{EA9999}0.33}     & \multicolumn{2}{c}{\cellcolor[HTML]{B6D7A8}1.10E-18} & \multicolumn{2}{c}{\cellcolor[HTML]{B6D7A8}0.001}    & \multicolumn{2}{c}{\cellcolor[HTML]{B6D7A8}2.88E-10}  \\
\textbf{Affect Words}          & \cellcolor[HTML]{B6D7A8}1.75E-09 & \multicolumn{2}{c}{\cellcolor[HTML]{EA9999}0.15}     & \multicolumn{2}{c}{\cellcolor[HTML]{B6D7A8}1.22E-33} & \multicolumn{2}{c}{\cellcolor[HTML]{EA9999}0.39}   & \multicolumn{2}{c}{\cellcolor[HTML]{EA9999}0.51}     & \multicolumn{2}{c}{\cellcolor[HTML]{B6D7A8}0.0001}   & \multicolumn{2}{c}{\cellcolor[HTML]{B6D7A8}7.51E-11} & \multicolumn{2}{c}{\cellcolor[HTML]{B6D7A8}0.001}     \\
\textbf{Social Words}          & \cellcolor[HTML]{B6D7A8}3.15E-07 & \multicolumn{2}{c}{\cellcolor[HTML]{EA9999}0.78}     & \multicolumn{2}{c}{\cellcolor[HTML]{B6D7A8}5.93E-17} & \multicolumn{2}{c}{\cellcolor[HTML]{EA9999}0.94}   & \multicolumn{2}{c}{\cellcolor[HTML]{EA9999}0.36}     & \multicolumn{2}{c}{\cellcolor[HTML]{B6D7A8}3.30E-19} & \multicolumn{2}{c}{\cellcolor[HTML]{B6D7A8}0.002}    & \multicolumn{2}{c}{\cellcolor[HTML]{B6D7A8}0.0006}    \\
\textbf{Cognitive Processes}   & \cellcolor[HTML]{B6D7A8}1.50E-07 & \multicolumn{2}{c}{\cellcolor[HTML]{B6D7A8}4.37E-05} & \multicolumn{2}{c}{\cellcolor[HTML]{B6D7A8}1.69E-21} & \multicolumn{2}{c}{\cellcolor[HTML]{EA9999}0.36}   & \multicolumn{2}{c}{\cellcolor[HTML]{EA9999}0.99}     & \multicolumn{2}{c}{\cellcolor[HTML]{B6D7A8}1.65E-09} & \multicolumn{2}{c}{\cellcolor[HTML]{B6D7A8}6.87E-05} & \multicolumn{2}{c}{\cellcolor[HTML]{B6D7A8}4.22E-09}  \\
\textbf{Perpetual Processes}   & \cellcolor[HTML]{EA9999}0.99     & \multicolumn{2}{c}{\cellcolor[HTML]{EA9999}0.96}     & \multicolumn{2}{c}{\cellcolor[HTML]{EA9999}0.11}     & \multicolumn{2}{c}{\cellcolor[HTML]{B6D7A8}0.01}   & \multicolumn{2}{c}{\cellcolor[HTML]{EA9999}0.06}     & \multicolumn{2}{c}{\cellcolor[HTML]{EA9999}0.98}     & \multicolumn{2}{c}{\cellcolor[HTML]{EA9999}0.79}     & \multicolumn{2}{c}{\cellcolor[HTML]{EA9999}0.31}      \\
\textbf{Biological Processes}  & \cellcolor[HTML]{EA9999}0.99     & \multicolumn{2}{c}{\cellcolor[HTML]{EA9999}0.28}     & \multicolumn{2}{c}{\cellcolor[HTML]{B6D7A8}6.66E-08} & \multicolumn{2}{c}{\cellcolor[HTML]{EA9999}0.999}  & \multicolumn{2}{c}{\cellcolor[HTML]{EA9999}0.54}     & \multicolumn{2}{c}{\cellcolor[HTML]{EA9999}0.59}     & \multicolumn{2}{c}{\cellcolor[HTML]{EA9999}1}        & \multicolumn{2}{c}{\cellcolor[HTML]{EA9999}0.46}      \\
\textbf{Core Drives and Needs} & \cellcolor[HTML]{EA9999}0.84     & \multicolumn{2}{c}{\cellcolor[HTML]{EA9999}0.09}     & \multicolumn{2}{c}{\cellcolor[HTML]{B6D7A8}5.98E-06} & \multicolumn{2}{c}{\cellcolor[HTML]{B6D7A8}0.0005} & \multicolumn{2}{c}{\cellcolor[HTML]{EA9999}0.38}     & \multicolumn{2}{c}{\cellcolor[HTML]{B6D7A8}0.03}     & \multicolumn{2}{c}{\cellcolor[HTML]{B6D7A8}0.003}    & \multicolumn{2}{c}{\cellcolor[HTML]{B6D7A8}0.0002}    \\
\textbf{Relativity}            & \cellcolor[HTML]{B6D7A8}3.86E-11 & \multicolumn{2}{c}{\cellcolor[HTML]{EA9999}0.63}     & \multicolumn{2}{c}{\cellcolor[HTML]{B6D7A8}0.005}    & \multicolumn{2}{c}{\cellcolor[HTML]{EA9999}0.39}   & \multicolumn{2}{c}{\cellcolor[HTML]{EA9999}0.07}     & \multicolumn{2}{c}{\cellcolor[HTML]{B6D7A8}0.002}    & \multicolumn{2}{c}{\cellcolor[HTML]{B6D7A8}0.009}    & \multicolumn{2}{c}{\cellcolor[HTML]{B6D7A8}0.038}     \\
\textbf{Informal Speech}       & \cellcolor[HTML]{B6D7A8}0.04     & \multicolumn{2}{c}{\cellcolor[HTML]{B6D7A8}3.68E-07} & \multicolumn{2}{c}{\cellcolor[HTML]{B6D7A8}0.001}    & \multicolumn{2}{c}{\cellcolor[HTML]{B6D7A8}0.01}   & \multicolumn{2}{c}{\cellcolor[HTML]{EA9999}0.12}     & \multicolumn{2}{c}{\cellcolor[HTML]{B6D7A8}0.0005}   & \multicolumn{2}{c}{\cellcolor[HTML]{EA9999}0.33}     & \multicolumn{2}{c}{\cellcolor[HTML]{B6D7A8}5.09E-13}  \\
\textbf{All Punctuation}       & \cellcolor[HTML]{B6D7A8}0.0003   & \multicolumn{2}{c}{\cellcolor[HTML]{EA9999}0.44}     & \multicolumn{2}{c}{\cellcolor[HTML]{B6D7A8}0.033}    & \multicolumn{2}{c}{\cellcolor[HTML]{EA9999}0.34}   & \multicolumn{2}{c}{\cellcolor[HTML]{EA9999}0.8}      & \multicolumn{2}{c}{\cellcolor[HTML]{EA9999}0.24}     & \multicolumn{2}{c}{\cellcolor[HTML]{EA9999}0.079}    & \multicolumn{2}{c}{\cellcolor[HTML]{B6D7A8}7.77E-16}  \\
\textbf{Personal Concerns}     & \cellcolor[HTML]{B6D7A8}3.75E-07 & \multicolumn{2}{c}{\cellcolor[HTML]{EA9999}0.2}      & \multicolumn{2}{c}{\cellcolor[HTML]{B6D7A8}8.28E-05} & \multicolumn{2}{c}{\cellcolor[HTML]{EA9999}0.2}    & \multicolumn{2}{c}{\cellcolor[HTML]{EA9999}0.22}     & \multicolumn{2}{c}{\cellcolor[HTML]{EA9999}0.18}     & \multicolumn{2}{c}{\cellcolor[HTML]{EA9999}0.057}    & \multicolumn{2}{c}{\cellcolor[HTML]{B6D7A8}0.0001}    \\
\textbf{Time Orientation}      & \cellcolor[HTML]{B6D7A8}0.001    & \multicolumn{2}{c}{\cellcolor[HTML]{EA9999}0.21}     & \multicolumn{2}{c}{\cellcolor[HTML]{B6D7A8}1.56E-05} & \multicolumn{2}{c}{\cellcolor[HTML]{EA9999}0.07}   & \multicolumn{2}{c}{\cellcolor[HTML]{EA9999}0.69}     & \multicolumn{2}{c}{\cellcolor[HTML]{B6D7A8}4.12E-05} & \multicolumn{2}{c}{\cellcolor[HTML]{EA9999}0.0791}   & \multicolumn{2}{c}{\cellcolor[HTML]{B6D7A8}0.024}     \\
\textbf{Grammar Other}         & \cellcolor[HTML]{B6D7A8}0.039    & \multicolumn{2}{c}{\cellcolor[HTML]{B6D7A8}0.002}    & \multicolumn{2}{c}{\cellcolor[HTML]{B6D7A8}0.01}     & \multicolumn{2}{c}{\cellcolor[HTML]{B6D7A8}0.03}   & \multicolumn{2}{c}{\cellcolor[HTML]{EA9999}0.86}     & \multicolumn{2}{c}{\cellcolor[HTML]{B6D7A8}0.014}    & \multicolumn{2}{c}{\cellcolor[HTML]{B6D7A8}0.03}     & \multicolumn{2}{c}{\cellcolor[HTML]{B6D7A8}0.01}      \\
\textbf{Language Metrics}      & \cellcolor[HTML]{B6D7A8}0.0005   & \multicolumn{2}{c}{\cellcolor[HTML]{EA9999}0.77}     & \multicolumn{2}{c}{\cellcolor[HTML]{B6D7A8}3.57E-14} & \multicolumn{2}{c}{\cellcolor[HTML]{EA9999}0.23}   & \multicolumn{2}{c}{\cellcolor[HTML]{EA9999}0.66}     & \multicolumn{2}{c}{\cellcolor[HTML]{B6D7A8}1.70E-06} & \multicolumn{2}{c}{\cellcolor[HTML]{EA9999}0.25}     & \multicolumn{2}{c}{\cellcolor[HTML]{B6D7A8}9.81E-14}  \\
\textbf{Summary Variable}      & \cellcolor[HTML]{B6D7A8}0        & \multicolumn{2}{c}{\cellcolor[HTML]{EA9999}0.87}     & \multicolumn{2}{c}{\cellcolor[HTML]{B6D7A8}2.75E-20} & \multicolumn{2}{c}{\cellcolor[HTML]{EA9999}0.36}   & \multicolumn{2}{c}{\cellcolor[HTML]{B6D7A8}1.95E-08} & \multicolumn{2}{c}{\cellcolor[HTML]{B6D7A8}0.012}    & \multicolumn{2}{c}{\cellcolor[HTML]{B6D7A8}0.006}    & \multicolumn{2}{c}{\cellcolor[HTML]{B6D7A8}1.13E-182}
\end{tabular}
}
\end{table}

\begin{table}[!htbp]

\caption{The table shows the significant difference of all \textcolor{blue}{LIWC} features between the reactions of non-rumour and \textcolor{blue}{rumour tweets}}
\label{tab:LIWCreactiontweets}
\resizebox{\columnwidth}{!} {%
\begin{tabular}{cclclclclclclclcl}
\textbf{}                      & \multicolumn{2}{c}{\textbf{Charlie}}                 & \multicolumn{2}{c}{\textbf{German}}                  & \multicolumn{2}{c}{\textbf{Sydney}}                  & \multicolumn{2}{c}{\textbf{Putin}}                   & \multicolumn{2}{c}{\textbf{Prince}}                  & \multicolumn{2}{c}{\textbf{Ottawa}}                  & \multicolumn{2}{c}{\textbf{Gurliit}}                 & \multicolumn{2}{c}{\textbf{Ferguson}}                \\
\textbf{WC}                    & \multicolumn{2}{c}{\cellcolor[HTML]{B6D7A8}3.36E-05} & \multicolumn{2}{c}{\cellcolor[HTML]{EA9999}0.06}     & \multicolumn{2}{c}{\cellcolor[HTML]{B6D7A8}0.04}     & \multicolumn{2}{c}{\cellcolor[HTML]{B6D7A8}0.048}    & \multicolumn{2}{c}{\cellcolor[HTML]{EA9999}0.5}      & \multicolumn{2}{c}{\cellcolor[HTML]{EA9999}0.34}     & \multicolumn{2}{c}{\cellcolor[HTML]{EA9999}0.93}     & \multicolumn{2}{c}{\cellcolor[HTML]{EA9999}0.55}     \\
\textbf{Function words}        & \multicolumn{2}{c}{\cellcolor[HTML]{B6D7A8}1.41E-32} & \multicolumn{2}{c}{\cellcolor[HTML]{EA9999}0.28}     & \multicolumn{2}{c}{\cellcolor[HTML]{B6D7A8}3.60E-28} & \multicolumn{2}{c}{\cellcolor[HTML]{B6D7A8}7.30E-01} & \multicolumn{2}{c}{\cellcolor[HTML]{EA9999}0.67}     & \multicolumn{2}{c}{\cellcolor[HTML]{B6D7A8}2.38E-05} & \multicolumn{2}{c}{\cellcolor[HTML]{EA9999}0.5}      & \multicolumn{2}{c}{\cellcolor[HTML]{B6D7A8}1.36E-22} \\
\textbf{Affect Words}          & \multicolumn{2}{c}{\cellcolor[HTML]{B6D7A8}3.01E-07} & \multicolumn{2}{c}{\cellcolor[HTML]{EA9999}0.052}    & \multicolumn{2}{c}{\cellcolor[HTML]{B6D7A8}2.17E-50} & \multicolumn{2}{c}{\cellcolor[HTML]{EA9999}0.64}     & \multicolumn{2}{c}{\cellcolor[HTML]{EA9999}0.79}     & \multicolumn{2}{c}{\cellcolor[HTML]{B6D7A8}0.014}    & \multicolumn{2}{c}{\cellcolor[HTML]{EA9999}0.52}     & \multicolumn{2}{c}{\cellcolor[HTML]{B6D7A8}1.40E-06} \\
\textbf{Social Words}          & \multicolumn{2}{c}{\cellcolor[HTML]{B6D7A8}2.42E-17} & \multicolumn{2}{c}{\cellcolor[HTML]{B6D7A8}0.0006}   & \multicolumn{2}{c}{\cellcolor[HTML]{B6D7A8}6.76E-26} & \multicolumn{2}{c}{\cellcolor[HTML]{EA9999}0.83}     & \multicolumn{2}{c}{\cellcolor[HTML]{EA9999}0.62}     & \multicolumn{2}{c}{\cellcolor[HTML]{B6D7A8}0.005}    & \multicolumn{2}{c}{\cellcolor[HTML]{EA9999}0.77}     & \multicolumn{2}{c}{\cellcolor[HTML]{B6D7A8}1.34E-06} \\
\textbf{Cognitive Processes}   & \multicolumn{2}{c}{\cellcolor[HTML]{B6D7A8}4.93E-17} & \multicolumn{2}{c}{\cellcolor[HTML]{B6D7A8}0.03}     & \multicolumn{2}{c}{\cellcolor[HTML]{B6D7A8}7.42E-27} & \multicolumn{2}{c}{\cellcolor[HTML]{EA9999}0.98}     & \multicolumn{2}{c}{\cellcolor[HTML]{EA9999}0.25}     & \multicolumn{2}{c}{\cellcolor[HTML]{B6D7A8}0.001}    & \multicolumn{2}{c}{\cellcolor[HTML]{EA9999}0.77}     & \multicolumn{2}{c}{\cellcolor[HTML]{B6D7A8}3.04E-28} \\
\textbf{Perpetual Processes}   & \multicolumn{2}{c}{\cellcolor[HTML]{B6D7A8}0.01}     & \multicolumn{2}{c}{\cellcolor[HTML]{EA9999}0.94}     & \multicolumn{2}{c}{\cellcolor[HTML]{EA9999}0.99}     & \multicolumn{2}{c}{\cellcolor[HTML]{EA9999}0.95}     & \multicolumn{2}{c}{\cellcolor[HTML]{EA9999}0.97}     & \multicolumn{2}{c}{\cellcolor[HTML]{B6D7A8}0.015}    & \multicolumn{2}{c}{\cellcolor[HTML]{EA9999}0.94}     & \multicolumn{2}{c}{\cellcolor[HTML]{B6D7A8}6.80E-08} \\
\textbf{Biological Processes}  & \multicolumn{2}{c}{\cellcolor[HTML]{EA9999}0.66}     & \multicolumn{2}{c}{\cellcolor[HTML]{EA9999}0.94}     & \multicolumn{2}{c}{\cellcolor[HTML]{EA9999}0.17}     & \multicolumn{2}{c}{\cellcolor[HTML]{EA9999}0.99}     & \multicolumn{2}{c}{\cellcolor[HTML]{EA9999}0.38}     & \multicolumn{2}{c}{\cellcolor[HTML]{EA9999}0.89}     & \multicolumn{2}{c}{\cellcolor[HTML]{EA9999}1}        & \multicolumn{2}{c}{\cellcolor[HTML]{EA9999}0.194}    \\
\textbf{Core Drives and Needs} & \multicolumn{2}{c}{\cellcolor[HTML]{B6D7A8}0.01}     & \multicolumn{2}{c}{\cellcolor[HTML]{B6D7A8}0.004}    & \multicolumn{2}{c}{\cellcolor[HTML]{B6D7A8}0.004}    & \multicolumn{2}{c}{\cellcolor[HTML]{EA9999}0.41}     & \multicolumn{2}{c}{\cellcolor[HTML]{EA9999}0.32}     & \multicolumn{2}{c}{\cellcolor[HTML]{B6D7A8}0.006}    & \multicolumn{2}{c}{\cellcolor[HTML]{EA9999}0.5}      & \multicolumn{2}{c}{\cellcolor[HTML]{EA9999}0.14}     \\
\textbf{Relativity}            & \multicolumn{2}{c}{\cellcolor[HTML]{B6D7A8}1.16E-42} & \multicolumn{2}{c}{\cellcolor[HTML]{EA9999}0.51}     & \multicolumn{2}{c}{\cellcolor[HTML]{B6D7A8}1.80E-10} & \multicolumn{2}{c}{\cellcolor[HTML]{EA9999}0.92}     & \multicolumn{2}{c}{\cellcolor[HTML]{EA9999}0.98}     & \multicolumn{2}{c}{\cellcolor[HTML]{EA9999}0.9}      & \multicolumn{2}{c}{\cellcolor[HTML]{EA9999}0.81}     & \multicolumn{2}{c}{\cellcolor[HTML]{B6D7A8}0.008}    \\
\textbf{Informal Speech}       & \multicolumn{2}{c}{\cellcolor[HTML]{B6D7A8}1.33E-05} & \multicolumn{2}{c}{\cellcolor[HTML]{EA9999}0.23}     & \multicolumn{2}{c}{\cellcolor[HTML]{B6D7A8}1.17E-05} & \multicolumn{2}{c}{\cellcolor[HTML]{B6D7A8}0.016}    & \multicolumn{2}{c}{\cellcolor[HTML]{EA9999}0.68}     & \multicolumn{2}{c}{\cellcolor[HTML]{EA9999}0.21}     & \multicolumn{2}{c}{\cellcolor[HTML]{EA9999}0.99}     & \multicolumn{2}{c}{\cellcolor[HTML]{B6D7A8}4.11E-19} \\
\textbf{All Punctuation}       & \multicolumn{2}{c}{\cellcolor[HTML]{EA9999}0.19}     & \multicolumn{2}{c}{\cellcolor[HTML]{EA9999}0.48}     & \multicolumn{2}{c}{\cellcolor[HTML]{B6D7A8}0.031}    & \multicolumn{2}{c}{\cellcolor[HTML]{EA9999}0.33}     & \multicolumn{2}{c}{\cellcolor[HTML]{EA9999}0.1}      & \multicolumn{2}{c}{\cellcolor[HTML]{EA9999}0.15}     & \multicolumn{2}{c}{\cellcolor[HTML]{EA9999}0.57}     & \multicolumn{2}{c}{\cellcolor[HTML]{B6D7A8}8.91E-10} \\
\textbf{Personal Concerns}     & \multicolumn{2}{c}{\cellcolor[HTML]{EA9999}0.25}     & \multicolumn{2}{c}{\cellcolor[HTML]{B6D7A8}0.001}    & \multicolumn{2}{c}{\cellcolor[HTML]{B6D7A8}0.002}    & \multicolumn{2}{c}{\cellcolor[HTML]{EA9999}0.99}     & \multicolumn{2}{c}{\cellcolor[HTML]{EA9999}0.99}     & \multicolumn{2}{c}{\cellcolor[HTML]{B6D7A8}0.0014}   & \multicolumn{2}{c}{\cellcolor[HTML]{EA9999}0.57}     & \multicolumn{2}{c}{\cellcolor[HTML]{EA9999}0.47}     \\
\textbf{Time Orientation}      & \multicolumn{2}{c}{\cellcolor[HTML]{B6D7A8}4.72E-15} & \multicolumn{2}{c}{\cellcolor[HTML]{B6D7A8}5.13E-05} & \multicolumn{2}{c}{\cellcolor[HTML]{B6D7A8}3.17E-10} & \multicolumn{2}{c}{\cellcolor[HTML]{EA9999}0.68}     & \multicolumn{2}{c}{\cellcolor[HTML]{EA9999}0.4}      & \multicolumn{2}{c}{\cellcolor[HTML]{B6D7A8}1.17E-10} & \multicolumn{2}{c}{\cellcolor[HTML]{EA9999}0.965}    & \multicolumn{2}{c}{\cellcolor[HTML]{B6D7A8}1.49E-08} \\
\textbf{Grammar Other}         & \multicolumn{2}{c}{\cellcolor[HTML]{B6D7A8}5.04E-08} & \multicolumn{2}{c}{\cellcolor[HTML]{B6D7A8}1.13E-08} & \multicolumn{2}{c}{\cellcolor[HTML]{B6D7A8}5.69E-14} & \multicolumn{2}{c}{\cellcolor[HTML]{EA9999}0.29}     & \multicolumn{2}{c}{\cellcolor[HTML]{EA9999}0.28}     & \multicolumn{2}{c}{\cellcolor[HTML]{B6D7A8}1.26E-05} & \multicolumn{2}{c}{\cellcolor[HTML]{B6D7A8}9.93E-07} & \multicolumn{2}{c}{\cellcolor[HTML]{B6D7A8}9.93E-07} \\
\textbf{Language Metrics}      & \multicolumn{2}{c}{\cellcolor[HTML]{B6D7A8}2.51E-12} & \multicolumn{2}{c}{\cellcolor[HTML]{B6D7A8}0.0005}   & \multicolumn{2}{c}{\cellcolor[HTML]{B6D7A8}1.22E-42} & \multicolumn{2}{c}{\cellcolor[HTML]{EA9999}0.27}     & \multicolumn{2}{c}{\cellcolor[HTML]{EA9999}0.09}     & \multicolumn{2}{c}{\cellcolor[HTML]{EA9999}0.132}    & \multicolumn{2}{c}{\cellcolor[HTML]{EA9999}0.75}     & \multicolumn{2}{c}{\cellcolor[HTML]{B6D7A8}7.91E-18} \\
\textbf{Summary Variable}      & \multicolumn{2}{c}{\cellcolor[HTML]{B6D7A8}0}        & \multicolumn{2}{c}{\cellcolor[HTML]{B6D7A8}2.47E-44} & \multicolumn{2}{c}{\cellcolor[HTML]{B6D7A8}0}        & \multicolumn{2}{c}{\cellcolor[HTML]{B6D7A8}5.09E-17} & \multicolumn{2}{c}{\cellcolor[HTML]{B6D7A8}8.36E-09} & \multicolumn{2}{c}{\cellcolor[HTML]{B6D7A8}3.66E-15} & \multicolumn{2}{c}{\cellcolor[HTML]{B6D7A8}0.004}    & \multicolumn{2}{c}{\cellcolor[HTML]{B6D7A8}0}       
\end{tabular}
}
\end{table}

\begin{table}[!htbp]
  \caption{The table shows the significant difference of all \textcolor{blue}{LIWC} features between the \textcolor{blue}{aggregated} non-rumour and aggregated rumour source (Src) and reaction tweets } %\fixme{\sout{Can we decrease the font of text in the table?}}
  \label{tab:LIWCaggregatedstats}
  \resizebox{4cm}{!}
  {%
\begin{tabular}{ccc}
\multicolumn{1}{l}{}           & \textbf{Src tweets}              & \textbf{Reaction}                \\
\textbf{WC}                    & \cellcolor[HTML]{B6D7A8}4.70E-05 & \cellcolor[HTML]{EA9999}9.99E-01 \\
\textbf{Function words}        & \cellcolor[HTML]{B6D7A8}4.70E-05 & \cellcolor[HTML]{EA9999}9.99E-01 \\
\textbf{Affect Words}          & \cellcolor[HTML]{B6D7A8}4.70E-05 & \cellcolor[HTML]{EA9999}9.99E-01 \\
\textbf{Social Words}          & \cellcolor[HTML]{B6D7A8}4.70E-05 & \cellcolor[HTML]{EA9999}9.99E-01 \\
\textbf{Cognitive Processes}   & \cellcolor[HTML]{B6D7A8}4.70E-05 & \cellcolor[HTML]{EA9999}9.99E-01 \\
\textbf{Perpetual Processes}   & \cellcolor[HTML]{B6D7A8}4.70E-05 & \cellcolor[HTML]{EA9999}9.99E-01 \\
\textbf{Biological Processes}  & \cellcolor[HTML]{B6D7A8}4.70E-05 & \cellcolor[HTML]{EA9999}9.99E-01 \\
\textbf{Core Drives and Needs} & \cellcolor[HTML]{B6D7A8}4.70E-05 & \cellcolor[HTML]{EA9999}9.99E-01 \\
\textbf{Relativity}            & \cellcolor[HTML]{B6D7A8}4.70E-05 & \cellcolor[HTML]{EA9999}9.99E-01 \\
\textbf{Informal Speech}       & \cellcolor[HTML]{B6D7A8}4.70E-05 & \cellcolor[HTML]{EA9999}9.99E-01 \\
\textbf{All Punctuation}       & \cellcolor[HTML]{B6D7A8}4.70E-05 & \cellcolor[HTML]{EA9999}9.99E-01 \\
\textbf{Personal Concerns}     & \cellcolor[HTML]{B6D7A8}3.75E-07 & \cellcolor[HTML]{EA9999}2.50E-01 \\
\textbf{Time Orientation}      & \cellcolor[HTML]{B6D7A8}4.20E-05 & \cellcolor[HTML]{EA9999}9.99E-01 \\
\textbf{Grammar Other}         & \cellcolor[HTML]{B6D7A8}0.03     & \cellcolor[HTML]{B6D7A8}5.05E-08 \\
\textbf{Language Metrics}      & \cellcolor[HTML]{EA9999}1        & \cellcolor[HTML]{EA9999}1        \\
\textbf{Summary Variable}      & \cellcolor[HTML]{B6D7A8}4.70E-05 & \cellcolor[HTML]{EA9999}1.00E+00
\end{tabular}}
\end{table}

%% file: Tables/Readability_stats.tex
\begin{table}[!htbp]

  \caption{The table shows the significant difference of all \textcolor{blue}{Readability} features between the non-rumour and rumour \textcolor{blue}{source tweets}}
   \label{tab:Readabilitysourcetweets}
     \resizebox{\columnwidth}{!}
  {%
\begin{tabular}{ccclclclclclclcl}
\textbf{}                     & \textbf{Charlie}                 & \multicolumn{2}{c}{\textbf{German}}                  & \multicolumn{2}{c}{\textbf{Sydney}}                  & \multicolumn{2}{c}{\textbf{Putin}}                   & \multicolumn{2}{c}{\textbf{Prince}}                & \multicolumn{2}{c}{\textbf{Ottawa}}                  & \multicolumn{2}{c}{\textbf{Gurliit}}                 & \multicolumn{2}{c}{\textbf{Ferguson}}                \\
\textbf{flesch\_score}        & \cellcolor[HTML]{EA9999}0.065    & \multicolumn{2}{c}{\cellcolor[HTML]{B6D7A8}4.35E-09} & \multicolumn{2}{c}{\cellcolor[HTML]{EA9999}0.11}     & \multicolumn{2}{c}{\cellcolor[HTML]{B6D7A8}0.005}    & \multicolumn{2}{c}{\cellcolor[HTML]{B6D7A8}0.0007} & \multicolumn{2}{c}{\cellcolor[HTML]{B6D7A8}1.61E-06} & \multicolumn{2}{c}{\cellcolor[HTML]{EA9999}0.378}    & \multicolumn{2}{c}{\cellcolor[HTML]{B6D7A8}1.00E-04} \\
\textbf{fleschkincaid\_score} & \cellcolor[HTML]{EA9999}1.90E-01 & \multicolumn{2}{c}{\cellcolor[HTML]{B6D7A8}6.42E-09} & \multicolumn{2}{c}{\cellcolor[HTML]{B6D7A8}1.00E-03} & \multicolumn{2}{c}{\cellcolor[HTML]{B6D7A8}0.039}    & \multicolumn{2}{c}{\cellcolor[HTML]{B6D7A8}0.035}  & \multicolumn{2}{c}{\cellcolor[HTML]{B6D7A8}1.00E-04} & \multicolumn{2}{c}{\cellcolor[HTML]{EA9999}0.31}     & \multicolumn{2}{c}{\cellcolor[HTML]{B6D7A8}3.10E-07} \\
\textbf{gunningfog\_score}    & \cellcolor[HTML]{B6D7A8}1.00E-02 & \multicolumn{2}{c}{\cellcolor[HTML]{EA9999}0.273}    & \multicolumn{2}{c}{\cellcolor[HTML]{B6D7A8}1.07E-09} & \multicolumn{2}{c}{\cellcolor[HTML]{EA9999}0.065}    & \multicolumn{2}{c}{\cellcolor[HTML]{EA9999}0.1074} & \multicolumn{2}{c}{\cellcolor[HTML]{B6D7A8}0.001}    & \multicolumn{2}{c}{\cellcolor[HTML]{B6D7A8}1.00E-02} & \multicolumn{2}{c}{\cellcolor[HTML]{B6D7A8}7.57E-05} \\
\textbf{smog\_score}          & \cellcolor[HTML]{EA9999}7.00E-01 & \multicolumn{2}{c}{\cellcolor[HTML]{EA9999}0.25}     & \multicolumn{2}{c}{\cellcolor[HTML]{B6D7A8}1.21E-07} & \multicolumn{2}{c}{\cellcolor[HTML]{EA9999}0.055}    & \multicolumn{2}{c}{\cellcolor[HTML]{EA9999}0.32}   & \multicolumn{2}{c}{\cellcolor[HTML]{EA9999}2.20E-01} & \multicolumn{2}{c}{\cellcolor[HTML]{B6D7A8}0.03}     & \multicolumn{2}{c}{\cellcolor[HTML]{EA9999}0.294}    \\
\textbf{dalechall\_score}     & \cellcolor[HTML]{B6D7A8}6.58E-06 & \multicolumn{2}{c}{\cellcolor[HTML]{B6D7A8}6.24E-06} & \multicolumn{2}{c}{\cellcolor[HTML]{B6D7A8}2.61E-05} & \multicolumn{2}{c}{\cellcolor[HTML]{B6D7A8}5.70E-07} & \multicolumn{2}{c}{\cellcolor[HTML]{EA9999}0.14}   & \multicolumn{2}{c}{\cellcolor[HTML]{B6D7A8}7.30E-08} & \multicolumn{2}{c}{\cellcolor[HTML]{EA9999}9.30E-01} & \multicolumn{2}{c}{\cellcolor[HTML]{EA9999}2.90E-01}
\end{tabular}
 }
\end{table}

\begin{table}[!htbp]
  \caption{The table shows the significant difference of all \textcolor{blue}{Readability} features between the \textcolor{blue}{reactions} of non-rumour and rumour tweets}
  \label{tab:Readabilityreactiontweets}
    \resizebox{\columnwidth}{!}
  {%
\begin{tabular}{cclclclclclclclcl}
\textbf{}                     & \multicolumn{2}{c}{\textbf{Charlie}}                 & \multicolumn{2}{c}{\textbf{German}}                  & \multicolumn{2}{c}{\textbf{Sydney}}                  & \multicolumn{2}{c}{\textbf{Putin}}                   & \multicolumn{2}{c}{\textbf{Prince}}              & \multicolumn{2}{c}{\textbf{Ottawa}}                  & \multicolumn{2}{c}{\textbf{Gurliit}}              & \multicolumn{2}{c}{\textbf{Ferguson}}                \\
\textbf{flesch\_score}        & \multicolumn{2}{c}{\cellcolor[HTML]{EA9999}5.49E-01} & \multicolumn{2}{c}{\cellcolor[HTML]{B6D7A8}0.005}    & \multicolumn{2}{c}{\cellcolor[HTML]{B6D7A8}0.01}     & \multicolumn{2}{c}{\cellcolor[HTML]{EA9999}0.86}     & \multicolumn{2}{c}{\cellcolor[HTML]{EA9999}0.66} & \multicolumn{2}{c}{\cellcolor[HTML]{B6D7A8}2.04E-10} & \multicolumn{2}{c}{\cellcolor[HTML]{EA9999}0.784} & \multicolumn{2}{c}{\cellcolor[HTML]{B6D7A8}0.009}    \\
\textbf{fleschkincaid\_score} & \multicolumn{2}{c}{\cellcolor[HTML]{B6D7A8}2.00E-03} & \multicolumn{2}{c}{\cellcolor[HTML]{B6D7A8}3.82E-06} & \multicolumn{2}{c}{\cellcolor[HTML]{B6D7A8}1.35E-08} & \multicolumn{2}{c}{\cellcolor[HTML]{EA9999}7.63E-01} & \multicolumn{2}{c}{\cellcolor[HTML]{EA9999}0.24} & \multicolumn{2}{c}{\cellcolor[HTML]{B6D7A8}2.00E-03} & \multicolumn{2}{c}{\cellcolor[HTML]{EA9999}0.69}  & \multicolumn{2}{c}{\cellcolor[HTML]{B6D7A8}3.66E-07} \\
\textbf{gunningfog\_score}    & \multicolumn{2}{c}{\cellcolor[HTML]{B6D7A8}3.94E-11} & \multicolumn{2}{c}{\cellcolor[HTML]{B6D7A8}0.00024}  & \multicolumn{2}{c}{\cellcolor[HTML]{B6D7A8}1.62E-10} & \multicolumn{2}{c}{\cellcolor[HTML]{EA9999}0.47}     & \multicolumn{2}{c}{\cellcolor[HTML]{EA9999}0.44} & \multicolumn{2}{c}{\cellcolor[HTML]{B6D7A8}0.02}     & \multicolumn{2}{c}{\cellcolor[HTML]{EA9999}0.797} & \multicolumn{2}{c}{\cellcolor[HTML]{B6D7A8}8.36E-05} \\
\textbf{smog\_score}          & \multicolumn{2}{c}{\cellcolor[HTML]{B6D7A8}3.27E-06} & \multicolumn{2}{c}{\cellcolor[HTML]{B6D7A8}0.01}     & \multicolumn{2}{c}{\cellcolor[HTML]{B6D7A8}2.58E-06} & \multicolumn{2}{c}{\cellcolor[HTML]{EA9999}0.53}     & \multicolumn{2}{c}{\cellcolor[HTML]{EA9999}0.91} & \multicolumn{2}{c}{\cellcolor[HTML]{B6D7A8}0.001}    & \multicolumn{2}{c}{\cellcolor[HTML]{EA9999}1}     & \multicolumn{2}{c}{\cellcolor[HTML]{EA9999}6.61E-01} \\
\textbf{dalechall\_score}     & \multicolumn{2}{c}{\cellcolor[HTML]{B6D7A8}6.00E-04} & \multicolumn{2}{c}{\cellcolor[HTML]{B6D7A8}0.0003}   & \multicolumn{2}{c}{\cellcolor[HTML]{B6D7A8}7.14E-05} & \multicolumn{2}{c}{\cellcolor[HTML]{EA9999}0.39}     & \multicolumn{2}{c}{\cellcolor[HTML]{EA9999}0.85} & \multicolumn{2}{c}{\cellcolor[HTML]{EA9999}0.06}     & \multicolumn{2}{c}{\cellcolor[HTML]{EA9999}0.17}  & \multicolumn{2}{c}{\cellcolor[HTML]{B6D7A8}4.40E-02}
\end{tabular}
}
\end{table}

%% file: Tables/SenticNet_stats.tex
\begin{table}[!htbp]
    \caption{The table shows the significant difference of all \textcolor{blue}{SenticNet} features between the \textcolor{blue}{aggregated} non-rumour and aggregated rumour source and reaction tweets}
    \label{tab:SenticAggregate}
    \resizebox{4cm}{!} {%
\begin{tabular}{ccc}
\multicolumn{1}{l}{}  & \textbf{Src tweets}              & \textbf{Reaction}                    \\
\textbf{Pleasantness} & \cellcolor[HTML]{B6D7A8}5.82E-05 & \cellcolor[HTML]{EA9999}0.9999999846 \\
\textbf{Attention}    & \cellcolor[HTML]{B6D7A8}5.82E-05 & \cellcolor[HTML]{EA9999}0.9999999846 \\
\textbf{Sensitivity}  & \cellcolor[HTML]{B6D7A8}4.64E-05 & \cellcolor[HTML]{EA9999}0.9999999846 \\
\textbf{Aptitude}     & \cellcolor[HTML]{B6D7A8}5.82E-05 & \cellcolor[HTML]{EA9999}0.9999999846 \\
\textbf{Polarity}     & \cellcolor[HTML]{B6D7A8}5.82E-05 & \cellcolor[HTML]{EA9999}0.9999999846
\end{tabular}
}
\end{table}

\begin{table}[!htbp]
\caption{The table shows the significant difference of all \textcolor{blue}{SenticNet} features between the  non-rumour and rumour \textcolor{blue}{source tweets}}
\label{tab:SenticNetsource}
  \resizebox{\columnwidth}{!}
  {%
\begin{tabular}{ccclclclclclclcl}
\textbf{}             & \textbf{Charlie}                 & \multicolumn{2}{c}{\textbf{German}}                  & \multicolumn{2}{c}{\textbf{Sydney}}                  & \multicolumn{2}{c}{\textbf{Putin}}                   & \multicolumn{2}{c}{\textbf{Prince}}               & \multicolumn{2}{c}{\textbf{Ottawa}}                  & \multicolumn{2}{c}{\textbf{Gurliit}}                 & \multicolumn{2}{c}{\textbf{Ferguson}}                \\
\textbf{pleasantness} & \cellcolor[HTML]{B6D7A8}1.84E-06 & \multicolumn{2}{c}{\cellcolor[HTML]{EA9999}4.30E-01} & \multicolumn{2}{c}{\cellcolor[HTML]{B6D7A8}3.51E-06} & \multicolumn{2}{c}{\cellcolor[HTML]{B6D7A8}3.37E-07} & \multicolumn{2}{c}{\cellcolor[HTML]{EA9999}0.975} & \multicolumn{2}{c}{\cellcolor[HTML]{B6D7A8}8.50E-09} & \multicolumn{2}{c}{\cellcolor[HTML]{EA9999}0.81}     & \multicolumn{2}{c}{\cellcolor[HTML]{EA9999}1.43E-01} \\
\textbf{attention}    & \cellcolor[HTML]{B6D7A8}4.40E-07 & \multicolumn{2}{c}{\cellcolor[HTML]{EA9999}3.40E-01} & \multicolumn{2}{c}{\cellcolor[HTML]{B6D7A8}2.00E-04} & \multicolumn{2}{c}{\cellcolor[HTML]{B6D7A8}0.0039}   & \multicolumn{2}{c}{\cellcolor[HTML]{B6D7A8}0.03}  & \multicolumn{2}{c}{\cellcolor[HTML]{B6D7A8}9.87E-08} & \multicolumn{2}{c}{\cellcolor[HTML]{EA9999}0.92}     & \multicolumn{2}{c}{\cellcolor[HTML]{EA9999}8.30E-01} \\
\textbf{sensitivity}  & \cellcolor[HTML]{B6D7A8}7.00E-04 & \multicolumn{2}{c}{\cellcolor[HTML]{EA9999}0.42}     & \multicolumn{2}{c}{\cellcolor[HTML]{B6D7A8}3.72E-05} & \multicolumn{2}{c}{\cellcolor[HTML]{B6D7A8}5.22E-07} & \multicolumn{2}{c}{\cellcolor[HTML]{EA9999}0.37}  & \multicolumn{2}{c}{\cellcolor[HTML]{EA9999}0.38}     & \multicolumn{2}{c}{\cellcolor[HTML]{EA9999}9.00E-01} & \multicolumn{2}{c}{\cellcolor[HTML]{B6D7A8}1.51E-07} \\
\textbf{aptitude}     & \cellcolor[HTML]{B6D7A8}1.77E-11 & \multicolumn{2}{c}{\cellcolor[HTML]{B6D7A8}0.03}     & \multicolumn{2}{c}{\cellcolor[HTML]{B6D7A8}1.10E-04} & \multicolumn{2}{c}{\cellcolor[HTML]{B6D7A8}0.0039}   & \multicolumn{2}{c}{\cellcolor[HTML]{EA9999}0.5}   & \multicolumn{2}{c}{\cellcolor[HTML]{B6D7A8}2.31E-08} & \multicolumn{2}{c}{\cellcolor[HTML]{EA9999}0.4431}   & \multicolumn{2}{c}{\cellcolor[HTML]{EA9999}0.26}     \\
\textbf{polarity}     & \cellcolor[HTML]{B6D7A8}5.54E-11 & \multicolumn{2}{c}{\cellcolor[HTML]{EA9999}5.50E-01} & \multicolumn{2}{c}{\cellcolor[HTML]{B6D7A8}2.19E-07} & \multicolumn{2}{c}{\cellcolor[HTML]{B6D7A8}2.31E-05} & \multicolumn{2}{c}{\cellcolor[HTML]{EA9999}0.93}  & \multicolumn{2}{c}{\cellcolor[HTML]{B6D7A8}2.11E-08} & \multicolumn{2}{c}{\cellcolor[HTML]{EA9999}3.70E-01} & \multicolumn{2}{c}{\cellcolor[HTML]{B6D7A8}3.40E-02}
\end{tabular}
}
\end{table}

\begin{table}[!htbp]
\caption{The table shows the significant difference of all \textcolor{blue}{SenticNet} features between the  \textcolor{blue}{reactions} of non-rumour and rumour tweets}
\label{tab:SenticNetreaction}
  \resizebox{\columnwidth}{!}
  {%
\begin{tabular}{cclclclclclclclcl}
\textbf{}             & \multicolumn{2}{c}{\textbf{Charlie}}                 & \multicolumn{2}{c}{\textbf{German}}                  & \multicolumn{2}{c}{\textbf{Sydney}}                  & \multicolumn{2}{c}{\textbf{Putin}}                   & \multicolumn{2}{c}{\textbf{Prince}}               & \multicolumn{2}{c}{\textbf{Ottawa}}                  & \multicolumn{2}{c}{\textbf{Gurliit}}             & \multicolumn{2}{c}{\textbf{Ferguson}}                \\
\textbf{pleasantness} & \multicolumn{2}{c}{\cellcolor[HTML]{B6D7A8}7.78E-10} & \multicolumn{2}{c}{\cellcolor[HTML]{B6D7A8}0.05}     & \multicolumn{2}{c}{\cellcolor[HTML]{B6D7A8}0.001}    & \multicolumn{2}{c}{\cellcolor[HTML]{EA9999}0.29}     & \multicolumn{2}{c}{\cellcolor[HTML]{EA9999}0.309} & \multicolumn{2}{c}{\cellcolor[HTML]{B6D7A8}5.33E-05} & \multicolumn{2}{c}{\cellcolor[HTML]{EA9999}0.82} & \multicolumn{2}{c}{\cellcolor[HTML]{EA9999}0.725}    \\
\textbf{attention}    & \multicolumn{2}{c}{\cellcolor[HTML]{B6D7A8}3.84E-09} & \multicolumn{2}{c}{\cellcolor[HTML]{EA9999}2.20E-01} & \multicolumn{2}{c}{\cellcolor[HTML]{B6D7A8}1.00E-04} & \multicolumn{2}{c}{\cellcolor[HTML]{EA9999}8.71E-01} & \multicolumn{2}{c}{\cellcolor[HTML]{EA9999}0.45}  & \multicolumn{2}{c}{\cellcolor[HTML]{B6D7A8}1.60E-04} & \multicolumn{2}{c}{\cellcolor[HTML]{EA9999}0.71} & \multicolumn{2}{c}{\cellcolor[HTML]{B6D7A8}1.10E-02} \\
\textbf{sensitivity}  & \multicolumn{2}{c}{\cellcolor[HTML]{B6D7A8}9.00E-03} & \multicolumn{2}{c}{\cellcolor[HTML]{EA9999}0.91}     & \multicolumn{2}{c}{\cellcolor[HTML]{B6D7A8}3.00E-03} & \multicolumn{2}{c}{\cellcolor[HTML]{EA9999}0.56}     & \multicolumn{2}{c}{\cellcolor[HTML]{EA9999}0.29}  & \multicolumn{2}{c}{\cellcolor[HTML]{B6D7A8}0.036}    & \multicolumn{2}{c}{\cellcolor[HTML]{EA9999}0.6}  & \multicolumn{2}{c}{\cellcolor[HTML]{B6D7A8}1.83E-07} \\
\textbf{aptitude}     & \multicolumn{2}{c}{\cellcolor[HTML]{B6D7A8}1.19E-24} & \multicolumn{2}{c}{\cellcolor[HTML]{EA9999}0.839}    & \multicolumn{2}{c}{\cellcolor[HTML]{B6D7A8}1.00E-03} & \multicolumn{2}{c}{\cellcolor[HTML]{EA9999}0.87}     & \multicolumn{2}{c}{\cellcolor[HTML]{EA9999}0.95}  & \multicolumn{2}{c}{\cellcolor[HTML]{B6D7A8}1.38E-06} & \multicolumn{2}{c}{\cellcolor[HTML]{EA9999}0.56} & \multicolumn{2}{c}{\cellcolor[HTML]{EA9999}7.00E-02} \\
\textbf{polarity}     & \multicolumn{2}{c}{\cellcolor[HTML]{B6D7A8}1.01E-20} & \multicolumn{2}{c}{\cellcolor[HTML]{EA9999}0.66}     & \multicolumn{2}{c}{\cellcolor[HTML]{B6D7A8}8.00E-04} & \multicolumn{2}{c}{\cellcolor[HTML]{EA9999}0.271}    & \multicolumn{2}{c}{\cellcolor[HTML]{EA9999}0.33}  & \multicolumn{2}{c}{\cellcolor[HTML]{B6D7A8}6.42E-05} & \multicolumn{2}{c}{\cellcolor[HTML]{EA9999}0.56} & \multicolumn{2}{c}{\cellcolor[HTML]{B6D7A8}9.00E-03}
\end{tabular}
}
\end{table}


%% file: Tables/Emotion_stats.tex
\begin{table}[!htbp]
  \caption{The table shows the percentage distribution of \textcolor{blue}{all the emotions} in the non-rumour and rumour classes. Src, and Re represents the source and reply tweets respectively.}
  \label{tab:Emotiondistribution}
    \resizebox{\columnwidth}{!}
  {%
\begin{tabular}{lcccc}
                  & \multicolumn{1}{l}{\textbf{Rumour Src}} & \multicolumn{1}{l}{\textbf{Non-rumour Src}} & \multicolumn{1}{l}{\textbf{Rumour Re}} & \multicolumn{1}{l}{\textbf{Non-rumour Re}} \\
\textbf{Anger}    & 7.83\%                                     & 18.52\%                                        & 15.64\%                                      & 20.60\%                                          \\
\textbf{Disgust}  & 1.59\%                                     & 3.13\%                                         & 3.74\%                                       & 4.49\%                                           \\
\textbf{Fear}     & 27.64\%                                    & 19.41\%                                        & 9.70\%                                       & 7.29\%                                           \\
\textbf{Joy}      & 4.15\%                                     & 8.20\%                                         & 6.55\%                                       & 9.03\%                                           \\
\textbf{Neutral}  & 19.30\%                                    & 28.11\%                                        & 35.85\%                                      & 36.24\%                                          \\
\textbf{Sadness}  & 27.14\%                                    & 15.16\%                                        & 12.99\%                                      & 9.33\%                                           \\
\textbf{Surprise} & 12.35\%                                    & 7.46\%                                         & 15.54\%                                      & 13.02\%                                         
\end{tabular}
}
\end{table}

%% file: Results.tex
\section{Prediction and Explainability}\label{sec:results}
\begin{table*}[!htbp]
\caption{Prediction Results of \textcolor{blue}{Random Forest} model on five events. Src, and Re represents the source and reply tweets respectively. Evaluation metrics are denoted as Acc (Accuracy), Pr (Precision), Rec (recall), and F1 (F1 Score).}
\label{tbl:prediction-results}
\begin{tabular}{c|c|c|c|c|c|c|c|c|c|c|c|c|c|c|c|c|}
\cline{2-17}
                          & \multicolumn{2}{c|}{\textbf{Charlie}} & \multicolumn{2}{c|}{\textbf{German}} & \multicolumn{2}{c|}{\textbf{Sydney}} & \multicolumn{2}{c|}{\textbf{Putin}} & \multicolumn{2}{c|}{\textbf{Prince}} & \multicolumn{2}{c|}{\textbf{Ottawa}} & \multicolumn{2}{c|}{\textbf{Gurlitt}} & \multicolumn{2}{c|}{\textbf{Ferguson}} \\ \cline{2-17} 
                          & \textbf{Src}              & \textbf{Re}              & \textbf{Src}          & \textbf{Re}           & \textbf{Src}          & \textbf{Re}           & \textbf{Src}          & \textbf{Re}          & \textbf{Src}           & \textbf{Re}          & \textbf{Src}          & \textbf{Re}           & \textbf{Src}           & \textbf{Re}           & \textbf{Src}           & \textbf{Re}            \\ \hline
\multicolumn{1}{|c|}{\textbf{Acc}} & 0.87             & 0.82            & 0.81         & 0.76         & 0.77         & 0.77         & 0.92         & 0.67        & 1             & 1           & 0.85         & 0.83         & 0.92          & 0.85         & 0.84           & 0.73          \\ \hline
\multicolumn{1}{|c|}{\textbf{Pr}}  & 0.87             & 0.83            & 0.81         & 0.76         & 0.77         & 0.77         & 0.93         & 0.67        & 1             & 1           & 0.85         & 0.85         & 0.92          & 0.85         & 0.84         & 0.74          \\ \hline
\multicolumn{1}{|c|}{\textbf{Rec}} & 0.87             & 0.82            & 0.81         & 0.76         & 0.77         & 0.77         & 0.92         & 0.67        & 1             & 1           & 0.85         & 0.85         & 0.92          & 0.85         & 0.84           & 0.73          \\ \hline
\multicolumn{1}{|c|}{\textbf{F1}}  & 0.87             & 0.76            & 0.81         & 0.76         & 0.77         & 0.77         & 0.91         & 0.67        & 1             & 1           & 0.85        & 0.85         & 0.92          & 0.85         & 0.84          & 0.65          \\ \hline
\end{tabular}
\end{table*}

\begin{figure*}[!htbp]
       \subfloat[Charlie hebdo (source tweets)  \label{subfig-1:shap_src}]{%
       \includegraphics[width=0.50\textwidth, height = 6.9cm]{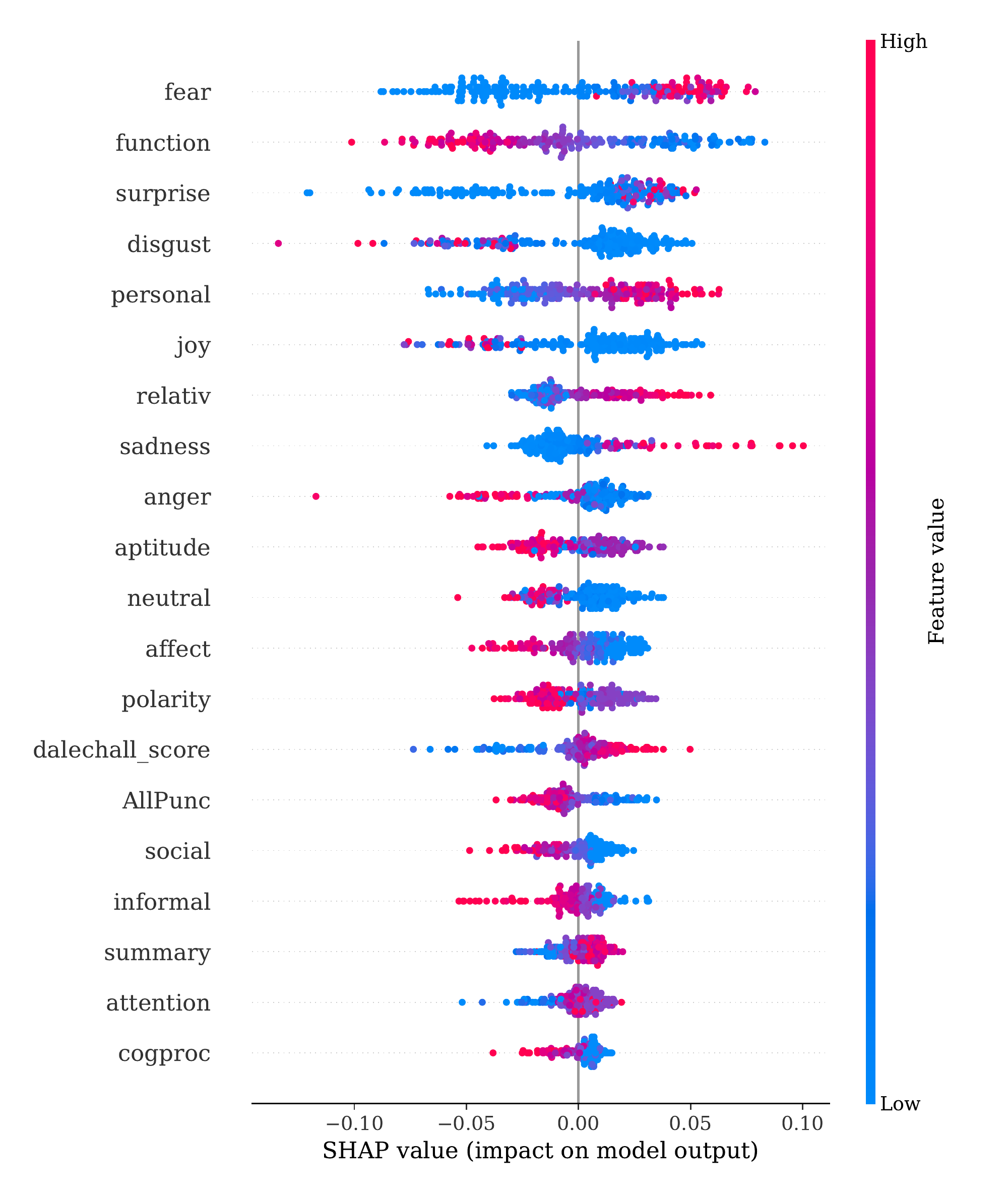}
     }
     \subfloat[Charlie hebdo (reply tweets) \label{subfig-2:shap_reply}]{%
       \includegraphics[width=0.50\textwidth, height = 6.9cm]{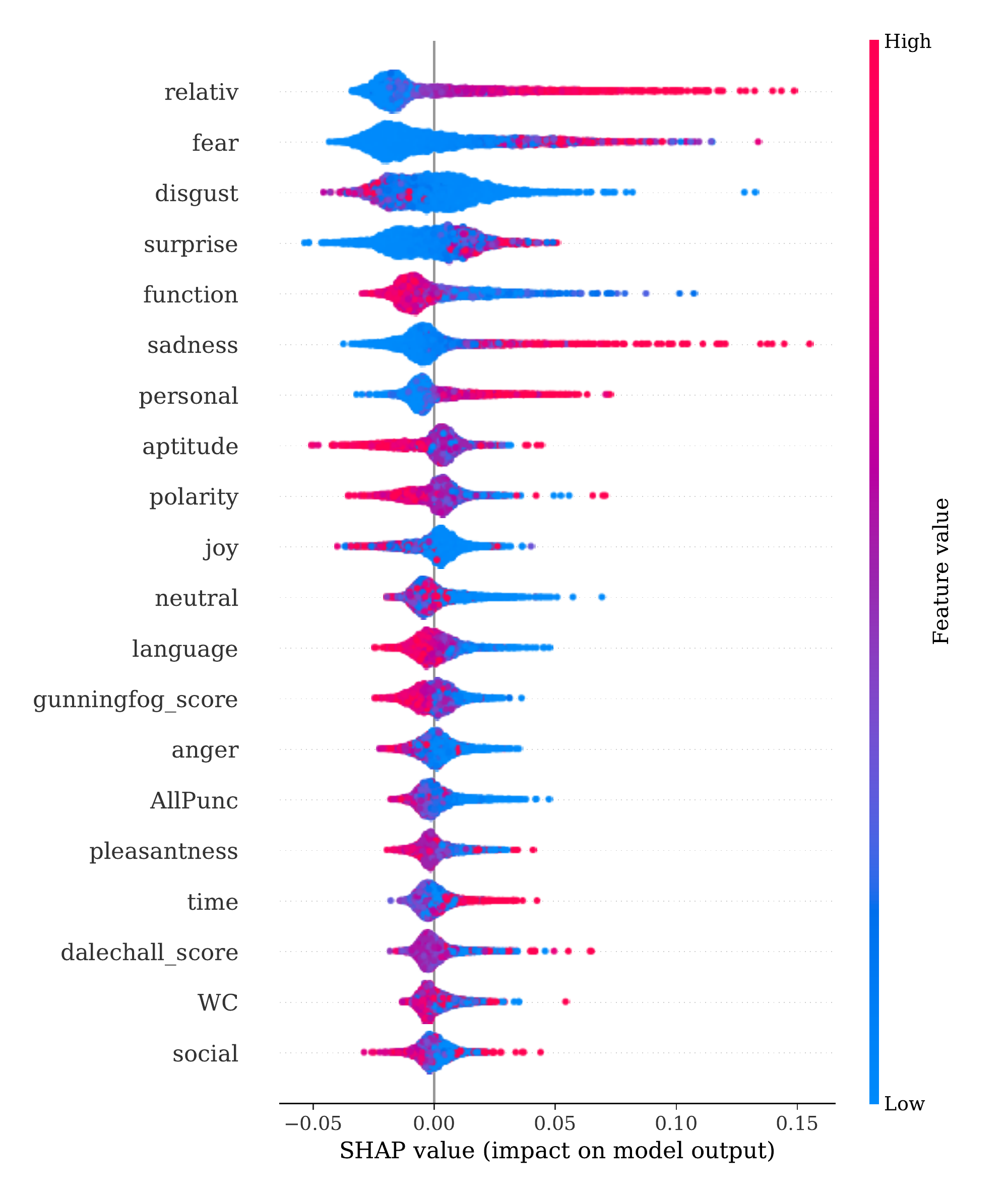}
     }
     \caption{SHAP tool to determine feature importance. The SHAP values and feature names are represented by the x- and y-axes, respectively. Each data point represents a single instance. The \textbf{\textcolor{red}{red}} color represents a higher value for the feature than its average value, while the \textbf{\textcolor{blue}{blue}} color denotes a lower value. A positive impact on the prediction is indicated by \textbf{ \textcolor{red}{red}} values on the right side of the x-axis, and vice versa. The features are listed in order of decreasing importance (best seen in color).}
     \label{fig:shap}
   \end{figure*}

%1. 30 ML models - CV, best performing, features, data split, events
%2. SHAP charlie event on RF src and reaction tweets.
In this section, %\fixme{combine this and next para?}
we investigate how effective are the four types of features (discussed in the previous section) distinguishing between the rumour and non-rumour categories (classes). We use these features as inputs to various Machine Learning (ML) models, both classical and ensemble-based, to categorise each tweet as rumour or non-rumour. Each event in the dataset is passed through 30 ML models individually.
Moreover, we train two separate models based on two types of inputs, that is, source tweets features and reply tweets features.
%Moreover, we train the models based on two types of inputs: source tweets features and reply tweets features separately. 
%\fixme{suggestion -- we train two separate models based on two types of inputs, that is, source tweets features and reply tweets features}
As previously stated, if a source tweet is classified as a rumour, all of their reply tweets are also classified as rumours and vice versa.
%\fixme{have we provided any explanation for this assumption in the previous section?}. 
This aids in determining whether the individual features of source and reply tweets are distinct enough to classify the tweet into one of two classes.

We split the source tweets (psycho-linguistic features) dataset into 80\% train and 20\% test set for each event. Further, we applied a ten-fold cross validation technique on the train set, which resulted in 10 train-fold sets. Following that, each ML model is trained on each train-fold set and then evaluated on the 20\% test set. We also apply the oversampling technique where the dataset is imbalanced. The same procedure is followed for reply tweets. We used default hyper-parameters of the ML models provided by scikit-learn\footnote{https://scikit-learn.org/stable/} package.

Due to space constraints, we only report the results of the best performing model, which is the Random Forest model, in Table \ref{tbl:prediction-results}. Each of the eight events is evaluated based on four metrics: Accuracy, Precision, Recall, and F1 Score.
The Table shows that the results of source tweets are better (\# of cases - 6) or equal (\# of cases - 2) to the results of reply tweets, implying that source tweets are more important to classify rumour and non-rumour tweets than reply tweets.
%The Table shows that the results of source tweets are better or equal to the results of reply tweets, implying that source tweets are more important to classify rumour and non-rumour tweets than reply tweets.

Next, the SHAP Explainability AI tool is utilized to evaluate the contributions of each feature in the classification task. Specifically, by computing the average marginal contributions of each feature, this tool assists in determining the significant features. Figure \ref{fig:shap} shows the significant features for the rumour class in descending order.
Due to space limits, we are only showing the SHAP plot of 
%\fixme{can we make "Charlie hebdo" italic?} 
\textit{Charlie hebdo} event for both source and reply tweets in Figures \ref{subfig-1:shap_src} and \ref{subfig-2:shap_reply} respectively.
%\fixme{Raj@Shakshi:Do we really need to write the next line or an we skip it?} We take into account all the attributes (features) in our classification task because each one is important for the prediction, as shown in this Figure (Despite the fact that the Figure only shows the top 20 features, all other features have positive SHAP values).
%It can be noted from the Figure \ref{subfig-1:shap_src} that the \textit{fear} is the highest contributing attribute in determining rumour class. Whereas \textit{relative} tops in Figure \ref{subfig-2:shap_reply} followed by the \textit{fear} attribute. 
%\fixme{suggestion for the previous line -- } 
It can be noted that for the source tweets, \textit{fear} is the highest contributing attribute, whereas for reply tweets \textit{relative} attributes the most.
We also illustrate the features that have a positive and negative impact on the class.
%\fixme{I am not sure if I got the meaning of the next line}
Particularly, in \textit{fear} attribute, as the red color samples, which are on the right side of the x-axis, are more than the blue color samples, this indicates a positive impact on the rumour class. This means that the higher the value of the \textit{fear} attribute, the better the chances of predicting the rumour class. This is intuitive as the \textit{fear} emotion is high in rumour tweets. 

The features: \textit{surprise, personal, sadness, dalechall\_score, relativ, summary, attention, drives, fleschkincaid\_score, grammar, WC, time, and percept} all have a positive impact on the rumour class, as seen in Figure \ref{subfig-1:shap_src}. We also noticed that the SHAP plot of reply tweets, Figure \ref{subfig-2:shap_reply} has the same but few different features than the SHAP plot of source tweets, Figure \ref{subfig-1:shap_src}: \textit{sensitivity, affect, informal, bio}, all of which have a positive influence on the rumour class.
The remaining attributes in both the Figures have a negative impact, meaning that as the value of these attributes decreases, the likelihood of correctly predicting rumour class increases.

%\fixme{We need to add something-- 
We noticed a similar trend in other terrorist-related or killing people events such as \textit{Sydney siege, Germanwings crash}. In comparison, in non-terrorist events such as \textit{Ferguson}, which is an event about protest, we observed that (not shown due to space limits), these events contains more positive influenced features with respect to rumour class than the \textit{Charlie} event, including \textit{language, fleshkincaid\_score, cogproc, surprise, sadness, personal, anger, gunningfog\_score, disgust, WC, function, drives, social, aptitude, bio, neutral, fear, time, polarity, attention, percept, smog\_score}.
%\sak{I could not think of any possible explanation for this line. If you have any suggestion, please add.} \sabur{
One possible explanation could be that the nature of the event was about speculation, future occurrence, and was driven by fear.
%The Ferguson event}
%\fixme{add-- (not shown due to space limit)} (both source and reply tweets), on the other hand, 
%cogproc, lang, disgust, function, surprise, fear, sadness, social, percept, neutral, flesh, gunning, time, apti, personal, -bio 

%\sabur{I think one paragraph comparing charlie hebdo with some other event is needed to explain the difference...}\sak{I am not sure. I think the main idea is to tell which features are important than the other features and not between events.}

\begin{comment}
\begin{figure}[!htbp]
    \centering
    \includegraphics[width=9cm, height = 7cm ]{Figures/shap_charlie_src_rumor.pdf}
    \caption{Feature Importance using SHAP tool. The x-axis and y-axis denote the SHAP values and features' names, respectively. Each data point refers to an instance of the dataset. The \textbf{\textcolor{red}{red}} color indicates a higher value for the feature than its average value, whereas the \textbf{\textcolor{blue}{blue}} color denotes a lower value.\textbf{ \textcolor{red}{Red}} values on the right side of the x-axis indicate a positive impact on the prediction and vice versa. Features are sorted in descending order (best seen in color).}\label{fig:shap}
\end{figure}
\end{comment}

%% file: Conclusion.tex
\section{Conclusions}\label{sec:conclusion}

The purpose of this research was to perform an in-depth analysis of the psycholinguistics side of the rumour task. We discovered a substantial difference between rumour and non-rumour psycho-linguistics source features, as well as between reply features.

We discovered that rumour source tweets used more past related words, prepositions and contain drives (motivation) related to reward, risk, and power. Similarly, non-rumour source tweets had more mean of features such as affective processes, cognitive processes (insight, causation, discrepancy, certainty, differentiation), present-related words, informal language, and had drives related to affiliation and achievement. The highest percentage of neutrality was found in non-rumor source tweets and non-rumor reactions, whereas rumour source tweets were driven by fear and grief, and their reactions invited anger and fear. Attention-related emotions (interest, anticipation, and vigilance) were more prevalent in rumour reactions, according to SenticNet. Readability has a considerable impact on the majority of events. We also explored the effectiveness of these features in predicting rumours. Specifically, we discovered that the ensemble-based, Random Forest model, for all events outperformed the other used models. As machine learning models are black-box in nature, we utilised the SHAP AI Explainability tool to look for the features that are more important than the other features. This helped in understanding which features contribute the most in classifying the tweets into one of two categories.

For future work, we plan to work in multiple directions. One possible future direction is to examine these features from a user-level perspective. This aids in the comprehension of rumour spreaders' human psychology. Another possible extension is to include more psycho-linguistic features, such as morphological features, referential cohesion, in our future study.

%Every category in LIWC contributed for the source tweets and their reactions. Some were more significant than the others and some contributed on in event specific scenarios.  Grammar other category played the most important role collectively to identify the reactions of rumour tweets. NRS and NRR had the highest percentage of neutrality, while RS was driven by fear and sadness and its reactions invited anger and fear. SenticNet: Attention related emotions (interest, anticipation and vigilance) were seen more in rumour reactions...... Also has more impact in the shap plots...... Polarity is also very important...Readability proved to be an important feature in ML classification ..... and impacting significant in majority of events.

%We also investigate the machine learning models to classify the tweets as rumour or non-rumour based on these psycho-linguistic features. Particularly, we found that ensemble-based, Random Forest model for all the events is performing the best among other 30 models. Since the machine learning models are black-box in nature, therefore, we investigate further using SHAP AI Explainability tool to look for the features that are more important from the classifier's perspective. 